\documentclass{article}

\PassOptionsToPackage{numbers, sort&compress}{natbib}

\usepackage[main, final]{neurips_2026}

\PassOptionsToPackage{dvipsnames,table}{xcolor}
\usepackage[utf8]{inputenc} 
\usepackage[T1]{fontenc}    
\usepackage[pagebackref]{hyperref}
\usepackage{backref}
\usepackage{url}            
\usepackage{booktabs}       
\usepackage{amsfonts}       
\usepackage{nicefrac}       
\usepackage{microtype}      

\usepackage{graphicx}
\usepackage{booktabs}
\usepackage{graphicx}
\usepackage{amsmath}
\usepackage{amssymb}
\usepackage{booktabs}
\usepackage{makecell}
\usepackage{multirow}
\usepackage{epigraph}
\usepackage{bm}
\usepackage{pifont}
\usepackage[many]{tcolorbox}
\usepackage{xspace}

\usepackage[capitalize]{cleveref}
\crefname{section}{Sec.}{Secs.}
\Crefname{section}{Section}{Sections}
\crefname{table}{Tab.}{Tabs.}
\Crefname{table}{Table}{Tables}

\usepackage{tikz}
\usepackage{pgfplots}
\pgfplotsset{compat=1.18}
\usetikzlibrary{positioning}
\usetikzlibrary{shapes.geometric}
\usetikzlibrary {shadows,calc}
\usetikzlibrary {arrows.meta}
\usetikzlibrary{calc}
\usepgfplotslibrary{groupplots}

\usepackage{subcaption}
\usepackage{import}
\usepackage{pifont}

\title{Balancing Frequencies and Pixels in Flow Matching}

\author{%
  Lucas Degeorge$^{1,2,3*}$ \quad
  Paul Couairon$^{1*}$ \quad
  Arijit Ghosh$^{1,3*}$ \\
  \textbf{Alexei A. Efros}$^{4}$ \quad
  \textbf{David Picard}$^{3\dagger}$ \quad
  \textbf{Vicky Kalogeiton}$^{1\dagger}$ \\[0.3cm]
  $^1$LIX, École Polytechnique, CNRS, IP Paris, France \quad
  $^2$AMIAD \\
  $^3$LIGM, École Nationale des Ponts et Chaussées, IP Paris, UGE, CNRS, France \quad
  $^4$UC Berkeley
}

\newcommand{\cmark}{\ding{51}}
\newcommand{\xmark}{\ding{55}}

\newcommand{\greyrule}{\arrayrulecolor{black!30}\midrule\arrayrulecolor{black}}
\newcommand{\greycmidrule}[1]{\arrayrulecolor{black!30}\cmidrule{#1}\arrayrulecolor{black}}

\definecolor{diffRedBg}{HTML}{FAECEC}
\definecolor{diffRedTxt}{HTML}{BA2A2A}
\definecolor{diffGreenBg}{HTML}{E8F5E8}
\definecolor{diffGreenTxt}{HTML}{278A3A}
\definecolor{diffComment}{HTML}{4A898A}
\definecolor{boxBorder}{HTML}{555555}

\definecolor{tableblue}{RGB}{110, 145, 220}
\definecolor{tablegreen}{RGB}{100, 145, 85}
\definecolor{tableyellow}{RGB}{255, 215, 0}
\definecolor{tablered}{RGB}{220, 100, 100}

\definecolor{pastelblue}{RGB}{140, 170, 240}
\definecolor{pastelgreen}{RGB}{130, 170, 110}
\definecolor{pastelyellow}{RGB}{247, 186, 100}
\definecolor{pastelred}{RGB}{255, 170, 170}

\newcommand{\floss}{\textcolor{black}{$\bm{f}$-loss}\xspace}
\newcommand{\vloss}{\textcolor{black}{$\bm{v}$-loss}\xspace}
\newcommand{\vfloss}{\textcolor{black}{$\bm{fv}$-loss}\xspace}

\newcommand{\flosstablecolor}{\textcolor{tableblue}{$\bm{f}$-loss}\xspace}
\newcommand{\vlosstablecolor}{\textcolor{tablegreen}{$\bm{v}$-loss}\xspace}
\newcommand{\vflosstablecolor}{\textcolor{tablered}{$\bm{fv}$-loss}\xspace}

\newcommand{\flosstablecolorttt}{\textcolor{tableblue}{$\bm{f}$-\texttt{loss}}\xspace}
\newcommand{\vlosstablecolorttt}{\textcolor{tablegreen}{$\bm{v}$-\texttt{loss}}\xspace}

\begin{document}

\maketitle

\vspace{-0.4cm}
\input{figures/teaser}

\begin{abstract}
Natural images follow a $1/f^2$ spectral distribution: most signal energy lies in the low spatial frequencies, while the perceptually important structures such as textures and edges occupy sparse high-frequency bands. Pixel-space reconstruction objectives, however, treat all spatial errors uniformly, causing low frequencies to dominate the optimization signal and delaying the learning of fine-scale details. 
In this work, we identify this objective-level spectral imbalance as a key inefficiency in training pixel-space flow models. 
To address it, we propose a Focal Log-Frequency Loss (\floss{}), a spectrally balanced objective that equalizes the learning signal across frequencies, emphasizing high-frequency components that are otherwise underrepresented in pixel-space objectives. 
Building on this, we introduce a simple training strategy that combines frequency and pixel supervision: we first emphasize frequency-domain learning early to capture all frequencies, and then transition to standard pixel-space $v$-loss for spatial refinement. This balancing mitigates the low-frequency bias of pixel losses and aligns the training signal with the evolving needs of the model. 
Our approach is conceptually simple, requires no architectural changes, and acts as a drop-in replacement for flow matching losses. Across multiple model scales, it accelerates convergence by up to $40\%$ while consistently improving FID and perceptual fidelity. 

\end{abstract}

\vspace{-0,2cm}

\section{Introduction}

Images admit two complementary representations: the spatial domain, which captures local structure, and the frequency domain, which captures global patterns across scales. Consider an image of a tiger. Its global shape is defined by smooth, large-scale variations, while the stripes -- thin, structured fine patterns -- are encoded in high frequencies. Both are essential: the overall silhouette identifies the object, while the fine patterns make it visually realistic. This duality between pixels and frequencies is fundamental to how images are structured and perceived \cite{geirhos2018imagenet,hermann2020origins,tartaglini2022developmentally}.

A key property of natural images is their spectral organization: their power spectrum approximately follows a $1/f^2$ distribution~\cite{ruderman1994statistics,torralba2003statistics}, meaning that most signal energy lies in low frequencies describing global structure, while high frequencies contain sparse but structured details. As a result, low-frequency components dominate the signal energy, even though they represent only part of the perceptual content of the image.
In fact, perceptual studies consistently show that high-frequency content plays a disproportionately important role in visual quality~\cite{Kayargadde96,Vansteenkiste06,Ferzli09,lieber2020naturalistic}. 
This suggests that image quality is highly sensitive to high-frequency components, such as edges and textures, despite their relatively low energy in the $1/f^2$ spectrum.
%

Motivated by this discrepancy, we 
introduce a frequency-domain objective that equalizes the training signal across the spectrum. Specifically, we propose a Focal Log-Frequency Loss (\floss{}), which applies focal weighting to logarithmically compressed spectral errors. In contrast to the $1/f^2$ distribution of natural images, this formulation increases the influence of high-frequency components that are otherwise underrepresented in pixel-space training~\cite{jit} (see Figure \ref{fig:raps}).

Building on this objective, we study its effect on training dynamics in flow matching models. We empirically show that pixel-space losses~\cite{1284395, Zhao2017LossFF, jit} are dominated by large low-frequency residuals, biasing early optimization. Emphasizing frequency-domain supervision counteracts this effect, enabling the model to capture structured patterns and fine-scale details early in training. We therefore adopt a simple two-stage strategy: we begin with \floss{} to learn spectral structure, and subsequently rely on the standard pixel \vloss \cite{jit} to refine spatial consistency. This simple strategy significantly improves both convergence speed and generation quality(see Figure~\ref{fig:teaser} and~\ref{fig:image_grid_512}). Importantly, our method acts as a drop-in replacement for standard flow matching objectives without requiring architectural changes~\cite{deco}, and achieves state-of-the-art performance both quantitatively (FID) and qualitatively for pixel-space class-conditional generation.

Our contributions are threefold:
\begin{itemize}
\item  We identify a fundamental mismatch between the spectral structure of natural images and pixel-space training objectives, showing that standard losses bias optimization toward low-frequency components despite the perceptual importance of high-frequency details;
\item  We introduce a spectrally balanced objective, the Focal Log-Frequency Loss (\floss{}), which equalizes the learning of frequencies and enables models to capture both global structure and fine-scale patterns; 
\item We demonstrate that emphasizing frequencies during training significantly improves optimization efficiency, while accelerating convergence and consistently improving generation quality across model scales, while requiring no architectural modifications.
\end{itemize}

\section{Related Work}

\paragraph{\textbf{Flow Matching and Pixel-Space Synthesis}} Denoising generative models including diffusion \cite{ddpm, ddim, dhariwal2021diffusion, karras2022elucidating, karras2024analyzing, song2020score, sohl2015deep, nichol2021improved} and flow matching frameworks \cite{lipmanflow, liuflow, albergobuilding}, typically frame the learning objective as a spatial-domain regression task, predicting either the noise ($\bm{\epsilon}$-prediction) or velocity field ($\bm{v}$-prediction). These perform quite well in recent methods operating in compressed pre-trained latent space \cite{rombach2022high, sdxl, sdv3, labs2025flux, ma2024sit, peebles2023scalable, chen2023pixart, chen2024pixart, zheng2026diffusion, xie2024sana}. 
However, when working directly on pixel space both $\bm{v}$-prediction and $\bm{\epsilon}$-prediction fail \cite{jit}. This is mainly due to the model's under-compute regime, which makes it difficult to separate the full-ranked noise components in its $\bm{v}$- or $\bm{\epsilon}$-prediction. Interestingly, natural data lie on a low-dimensional manifold \cite{Chapelle2006SemiSupervised}, as such, a direct denoised clean data prediction ($\bm{x}$-prediction) offers the best alternative \cite{jit}. This opens the door to adapt architectural modifications \cite{pixeldit, deco, hoogeboom2025simpler, wang2026pixnerd, chen2025dip, chen2025pixelflow} as well as additional loss terms \cite{pixelgen, lei2026there} for performance gains. 

In this work, we look into the main training objective of such pixel space models without any architectural modifications and auxiliary losses and find that a frequency rebalancing loss helps in faster convergence and improved performance.

\paragraph{Spectral Bias and Frequency-Aware Generative Modeling} Neural networks inherently exhibit a spectral bias, learning low-frequency structural components significantly faster than fine-grained, high-frequency details \cite{spectralbias, xu2019training}. Across generative paradigms, this bias yields models that excel at synthesizing global structure but struggle to preserve sharp textures \cite{karras2021alias, rombach2022high}, often resulting in severe high-frequency artifacts \cite{Schwarz2021NEURIPS}. To mitigate these limitations, prior works have explored various interventions. Some approaches balance frequency components post-hoc during sampling \cite{si2023freeu, 2024fouriscale}. Other works utilize temporal cascading to explicitly decouple frequencies across denoising timesteps \cite{chen2025pixelflow, teng2023relay}. 
More recently, researchers have introduced structural modifications like multi-scale learning strategies \cite{NIPS2015_aa169b49, karras2018progressive, kang2023scaling, gu2024matryoshka}, dedicated frequency branches \cite{yoon2024frag, ren2026frequencyawareflowmatchinghighquality}, and explicit multi-scale or latent frequency decoupling within the architecture itself \cite{wang2025ddt, deco, ffl}. 

While these methods underscore the critical importance of frequency balancing, they rely on complex denoising schedules, post-hoc adjustments, or complex architectural changes. In contrast, our work targets the fundamental optimization dynamics. We directly correct the spectral bias during training without requiring any architectural modifications.


\section{Method}
\label{sec:method}

\subsection{Background}

\paragraph{\textbf{Flow Matching Models}}~\cite{albergobuilding,liuflow,lipmanflow} define a regression objective to learn a velocity field $\bm{v}_{\theta}(\bm{x}_t, t)$ that transports noise $\bm{x}_0 \sim p_0$ to data $\bm{x}_1 \sim p_1$ via the ODE $\frac{dx_t}{dt} = \bm{v}_\theta(\bm{x}_t, t) = \bm{v}_\theta$. By employing a linear interpolant $\bm{x}_t = (1 - t)\,\boldsymbol{\epsilon} + t\,\bm{x}_1$, the model can be trained to directly predict the constant ground-truth velocity $\bm{v}^* = \bm{x}_1 - \boldsymbol{\epsilon}$. This approach has been widely adopted by state-of-the-art architectures~\cite{sdv3, ma2024sit} for its stability: 
\begin{equation}
\label{eq:vloss}
    \mathcal{L}_{\bm{v}} = \mathbb{E}\lVert \bm{v}_{\theta}- \bm{v}^*\rVert^2\, .
\end{equation}

\paragraph{\textbf{Prediction Parameterization}} The choice of prediction target is significantly influenced by the dimensionality of the data. While $\bm{v}$-prediction is highly effective within compressed latent~\cite{sdv3,labs2025flux1kontextflowmatching}, it often suffers from catastrophic failure when applied to high-dimensional pixel spaces~\cite{jit}. To address this, the objective is reparameterized to focus the model on $\bm{x}$-prediction rather than direct velocity regression. Under this framework, the ground-truth velocity is reformulated as: $\bm{v} = \bm{x}_1 - \boldsymbol{\epsilon} = \frac{\bm{x}_1 - \bm{x}_t}{1 - t} $ while $\bm{v}_{\theta} = \frac{\bm{x}_{\theta} - \bm{x}_t}{1 -t}$. Consequently, the flow matching velocity loss in Equation \ref{eq:vloss}  can be reformulated as a weighted $x$-regression loss, which we call \vloss:

\begin{equation}
    \mathcal{L}_{\bm{v}} = \mathbb{E}\left[\frac{1}{(1-t)^2}\lVert \bm{x}_{\theta}(\bm{x}_t, t) - \bm{x}_1\rVert^2\right] 
\end{equation}

By shifting to $\bm{x}$-regression, the model is tasked with projecting directly onto the data manifold, a signal that is significantly more tractable than regressing high-variance velocity fields across the high-dimensional ambient space. This better aligns with the model’s capacity by focusing the learning process on recovering low dimensional geometric structure rather than modeling high-dimensional temporal transitions.

\subsection{Spectral Diagnosis of Pixel-Space Flow Models}
\label{sec:spectral_diagnosis}

Neural networks are known to exhibit a spectral bias: they learn low-frequency components of a target function significantly faster than high-frequency ones, with the learning speed at frequency $k$ decaying approximately as $1/k$. Prior works~\cite{spectralbias, xu2019training} have studied theoretically the bias in the supervised learning setting. We investigate this question empirically through two complementary experiments: a quantitative analysis of the frequency signature of a fully trained model with \vloss, and a controlled toy experiment that tracks spectral learning dynamics over the course of training.

\begin{figure}[htbp]
    \centering
    
    \begin{subfigure}[c]{0.47\textwidth}
        \centering
        \input{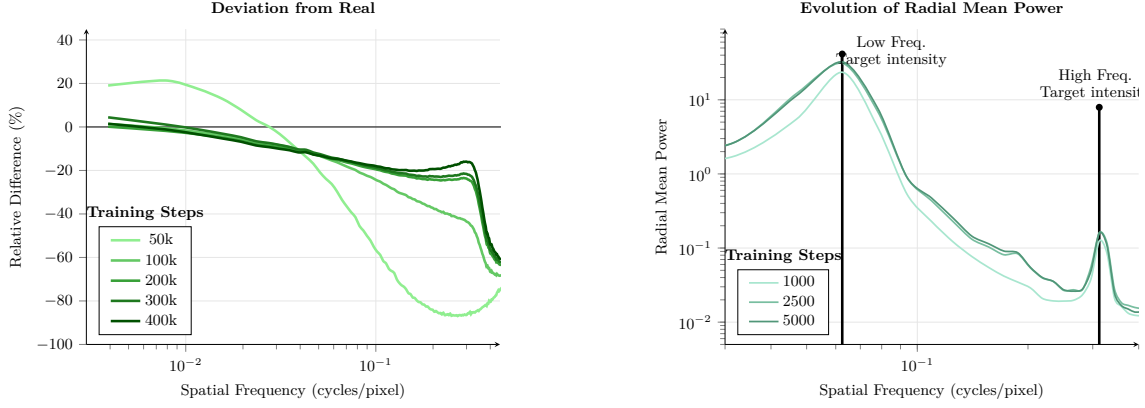} 
        \vspace{-12pt}
        \caption{Relative deviation of the generated image power spectrum from real images throughout training. \vloss overestimates low frequencies and underestimates high frequencies.}
        \label{fig:raps}
    \end{subfigure}
    \hfill
    \begin{subfigure}[c]{0.5\textwidth}
        \centering

\definecolor{lightgreen}{RGB}{168,230,207}
\definecolor{mediumgreen}{RGB}{120,190,160}
\definecolor{darkgreen}{RGB}{80,150,120}

\begin{tikzpicture}[scale=0.65, transform shape]
\begin{loglogaxis}[
    width=10cm,
    height=8cm,
    title={Evolution of Radial Mean Power},
    xlabel={Spatial Frequency (cycles/pixel)},
    ylabel={Radial Mean Power},
    xmin=0.03, xmax=0.4,
    ymin=0.005, ymax=90,
    xtick={0.001, 0.01, 0.1, 1},
    xticklabels={$10^{-3}$, $10^{-2}$, $10^{-1}$, $10^0$},
    grid=major,                         
    major grid style={thin, gray!20},   
    minor tick num=9,                   
    grid style={gray!30, line width=0.3pt},
    axis lines=left,
    tick align=outside,
    legend pos=south west,
    legend style={
    font=\small, 
    draw=black,
    inner xsep=6pt,
    label={[font=\footnotesize\bfseries]above:Training Steps},
},
    title style={font=\small\bfseries},
    label style={font=\small},
    tick label style={font=\small},
    clip=true,
]

\addplot[black, line width=1.5pt, no marks, forget plot] coordinates {
    (0.06250, 0.005) (0.06250, 41.47624)
};

\addplot[black, line width=1.5pt, no marks, forget plot] coordinates {
    (0.31250, 0.005) (0.31250, 7.92396)
};


\addplot[
    color=lightgreen,
    line width=1pt,
    smooth,
    no marks,
] coordinates {
(0.01562,1.20015) (0.03125,1.69209) (0.04688,5.45738) (0.06250,23.52914) (0.07812,4.20606) 
(0.09375,0.53290) (0.10938,0.22798) (0.12500,0.12153) (0.14062,0.07702) (0.15625,0.05480) 
(0.17188,0.04189) (0.18750,0.03475) (0.20312,0.02894) (0.21875,0.02094) (0.23438,0.01933) 
(0.25000,0.01925) (0.26562,0.01970) (0.28125,0.02336) (0.29688,0.03857) (0.31250,0.12443) 
(0.32812,0.08450) (0.34375,0.02112) (0.35938,0.01625) (0.37500,0.01317) (0.39062,0.01237) 
(0.40625,0.01219) (0.42188,0.01219) (0.43750,0.01268) (0.45312,0.01082) (0.46875,0.01105) 
(0.48438,0.01158) (0.50000,0.01113) (0.51562,0.00949) (0.53125,0.00878) (0.54688,0.00931) 
(0.56250,0.01015) (0.57812,0.01097) (0.59375,0.00782) (0.60938,0.00869) (0.62500,0.00973) 
(0.64062,0.00910) (0.65625,0.01048) (0.67188,0.00780) (0.68750,0.00843) (0.70312,0.01766)
};
\addlegendentry{1000}

\addplot[
    color=mediumgreen,
    line width=1pt,
    smooth,
    no marks,
] coordinates {
(0.01562,1.72813) (0.03125,2.57019) (0.04688,12.09902) (0.06250,31.16219) (0.07812,6.89738) 
(0.09375,0.89335) (0.10938,0.43412) (0.12500,0.22851) (0.14062,0.13192) (0.15625,0.10341) 
(0.17188,0.08120) (0.18750,0.08381) (0.20312,0.05388) (0.21875,0.03957) (0.23438,0.03537) 
(0.25000,0.02685) (0.26562,0.02734) (0.28125,0.02745) (0.29688,0.05588) (0.31250,0.16019) 
(0.32812,0.11591) (0.34375,0.02804) (0.35938,0.01823) (0.37500,0.01682) (0.39062,0.01577) 
(0.40625,0.01526) (0.42188,0.01481) (0.43750,0.01461) (0.45312,0.01537) (0.46875,0.01240) 
(0.48438,0.01274) (0.50000,0.01051) (0.51562,0.00818) (0.53125,0.00813) (0.54688,0.00821) 
(0.56250,0.00825) (0.57812,0.00941) (0.59375,0.00847) (0.60938,0.00774) (0.62500,0.00831) 
(0.64062,0.00838) (0.65625,0.00882) (0.67188,0.00820) (0.68750,0.00980) (0.70312,0.00782)
};
\addlegendentry{2500}

\addplot[
    color=darkgreen,
    line width=1pt,
    smooth,
    no marks,
] coordinates {
(0.01562,1.39432) (0.03125,2.56867) (0.04688,11.30742) (0.06250,32.46872) (0.07812,7.63066) 
(0.09375,0.90157) (0.10938,0.47257) (0.12500,0.25471) (0.14062,0.14500) (0.15625,0.11428) 
(0.17188,0.09109) (0.18750,0.08658) (0.20312,0.05473) (0.21875,0.04002) (0.23438,0.03548) 
(0.25000,0.02737) (0.26562,0.02619) (0.28125,0.02790) (0.29688,0.04947) (0.31250,0.15939) 
(0.32812,0.10751) (0.34375,0.02465) (0.35938,0.01669) (0.37500,0.01500) (0.39062,0.01366) 
(0.40625,0.01384) (0.42188,0.01404) (0.43750,0.01362) (0.45312,0.01368) (0.46875,0.01115) 
(0.48438,0.01112) (0.50000,0.00973) (0.51562,0.00709) (0.53125,0.00627) (0.54688,0.00663) 
(0.56250,0.00654) (0.57812,0.00657) (0.59375,0.00644) (0.60938,0.00604) (0.62500,0.00648) 
(0.64062,0.00630) (0.65625,0.00668) (0.67188,0.00523) (0.68750,0.00785) (0.70312,0.00658)
};
\addlegendentry{5000}

\node[font=\footnotesize, align=center] at (axis cs:0.085, 45) {Low Freq.  \\ Target intensity};
\node[font=\footnotesize, align=center] at (axis cs:0.305, 16) {High Freq. \\ Target intensity};

\addplot[black, only marks, mark=*, mark size=1.5pt, mark options={line width=1pt}] coordinates {
    (0.06250, 41.47624)
};
\addplot[black, only marks, mark=*, mark size=1.5pt, mark options={line width=1pt}] coordinates {
    (0.31250, 7.92396)
};
\end{loglogaxis}
\end{tikzpicture}
        \caption{Spectral evolution of a MLP trained with \vloss on a two-frequency toy signal. The low-frequency pattern (left) is learned rapidly; the high-frequency pattern (right) remains absent throughout training.}
        \label{fig:toy_example}
    \end{subfigure}
    
    \caption{\textbf{Spectral bias induced by \vloss}: (a) Low frequencies are overrepresented in the radial power spectrum of generated images, (b) High frequencies not learned as well as low frequencies on a simple toy experiment with only 2 active frequencies.}
    \label{fig:spectral_biais}
\end{figure}

\paragraph{Frequency signature of \vloss models} To directly measure the spectral distortion induced by \vloss, we compute the radially averaged power spectrum of images generated by a model trained with \vloss, and compare it to the power spectrum of real ImageNet images. Figure~\ref{fig:raps} reports the relative difference at each spatial frequency. \vloss model overestimates by 20\% low and mid frequencies (below $10^{-1}$ cycles/pixel). On the contrary, high frequencies (above $8\times10^{-1}$ cycles/pixel) are underestimated, with the deficit growing toward the Nyquist limit and reaching nearly -60\%.

\paragraph{Spectral learning dynamics} To isolate the temporal dynamics of this bias, we train a small MLP to generate synthetic images composed of exactly two frequency components: one low and one high. We track the Fourier spectrum of the generated images throughout training. As shown in Figure~\ref{fig:toy_example}, the model rapidly learns the low-frequency pattern but struggles to learn the high-frequency one. This result confirms that \vloss imposes an implicit preference that puts more emphasis on low frequencies throughout training, making it slow to learn high frequencies associated with fine grain textures. Similarly, in Figure~\ref{fig:raps}, the model manages to recover the right amount of low frequencies at the end of training, but is still missing some high frequencies. This suggests that by focusing too much on the low frequencies during early training, the capacity of the model to generate high frequencies is irremediably lost and cannot be recovered at later stages.

\subsection{A Joint Pixel-Frequency Objective}

As mentioned in Section~\ref{sec:spectral_diagnosis}, supervising only pixel-space  residuals creates a systematic spectral imbalance that persists throughout training.  This motivates a training objective that operates on both domains jointly: pixel-space  supervision to ensure spatial precision, and frequency-space supervision to ensure  spectral balance.

\paragraph{Frequency-domain component}

To ensure such spectral balance, we introduce \floss, a frequency-domain reconstruction objective that  reweights the learning signal across the spectral hierarchy:

\begin{equation}
    \mathcal{L}_{\bm{f}} = \sum_{u,v} \frac{1}{(1-t)^2} \cdot \frac{e_{u,v}}{\displaystyle\max_{u',v'} e_{u',v'}} \cdot \log\!\left(1 + \left|\mathcal{F}_\text{pred}(u,v) - \mathcal{F}_\text{target}(u,v)\right|\right)
\end{equation}

where $\mathcal{F}$ denotes the 2D Discrete Fourier Transform and $e_{u,v} = |\mathcal{F}_\text{pred}(u,v) - \mathcal{F}_\text{target}(u,v)|$ is the per-frequency residual.

The three components each serve a distinct role. First, the \textbf{timestep weighting} $\frac{1}{(1-t)^2}$ is inherited from \vloss and preserved unchanged, ensuring consistent timesteps dynamics. Then, the \textbf{adaptive focal weight} $\frac{e_{u,v}}{\displaystyle\max_{u',v'} e_{u',v'}}$  normalizes the per-frequency residuals by the maximum residual across the spectrum at each training step. This is computed with a stop-gradient and acts as a purely data-dependent coefficient: it continuously normalizes each sample to avoid wide variations in magnitude between samples. Finally, the \textbf{logarithmic compression} $\log(1 + e_{u,v})$ prevents any single frequency from monopolizing the loss, and can be interpreted as a continuous generalization of Laplacian pyramid decomposition: just as the pyramid assigns equal structural importance to each octave, the logarithm linearizes the spectral hierarchy, granting every doubling of frequency equal weight in the total loss (see Appendix~\ref{app:laplacian}).

\paragraph{Joint objective}

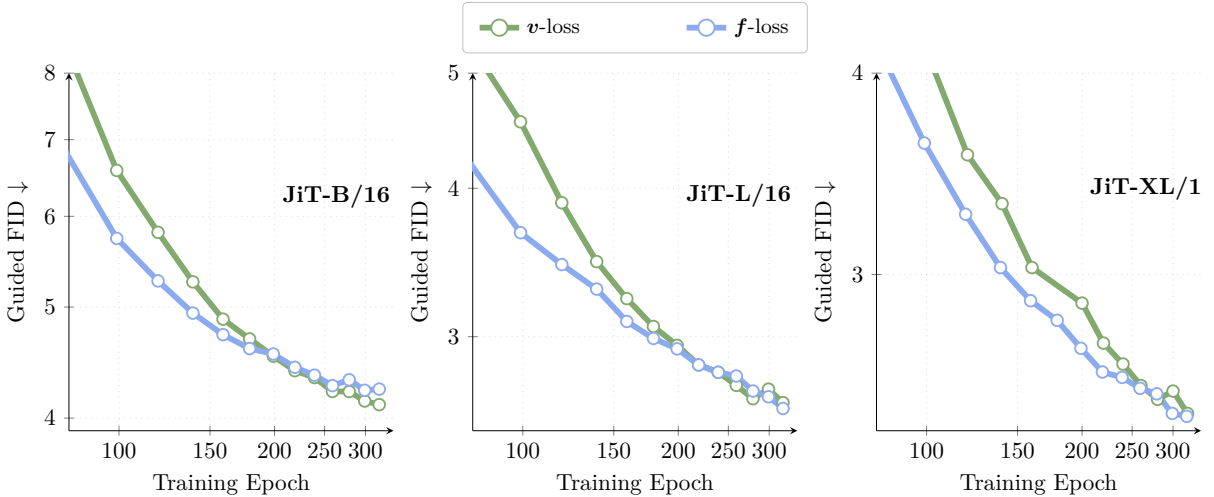
\begin{figure}[t]
    \centering
    \resizebox{\textwidth}{!}{
        \definecolor{vblue}{RGB}{27, 158, 119}
\definecolor{fflred}{RGB}{217, 95, 2}

\pgfplotsset{group/every plot/.style={}}

\begin{tikzpicture}
    \begin{groupplot}[
        group style={
            group size=3 by 1, 
            horizontal sep=1.2cm, 
            vertical sep=1.5cm,   
        },
        width=6.5cm,
        height=7cm,
        axis lines=left, 
        grid=major,
        grid style={dotted, gray!30},
        xlabel={Training Epoch},
        legend style={draw=none, fill=none},
        every axis plot/.append style={
        line width=2.5pt,
        mark size=2.5pt,
        mark options={fill=white, thick} 
        }
    ]

    \nextgroupplot[
        ylabel={Guided FID $\downarrow$},
        xmode=log,
        ymode=log,
        xmin=80, xmax=340,
        xtick={50, 100, 150, 200, 250, 300},
        xticklabels={50, 100, 150, 200, 250, 300},
        ymin=3.9, ymax=8,
        ytick={4, 5, 6, 7, 8, 10, 12, 14},
        log ticks with fixed point,
    ]
        \addplot[color=pastelgreen, mark=*] coordinates {
            (59, 13.42) (79, 8.44) (99, 6.58) (119, 5.81) (139, 5.26) (159, 4.88) (179, 4.69) (199, 4.53) (219, 4.40) (239, 4.34) (259, 4.22) (279, 4.22) (299, 4.14) (319, 4.11)
        };
    
        \addplot[color=pastelblue, mark=*] coordinates {
            (59, 9.80) (79, 6.82) (99, 5.74) (119, 5.27) (139, 4.94) (159, 4.73) (179, 4.60) (199, 4.55) (219, 4.43) (239, 4.36) (259, 4.27) (279, 4.32) (299, 4.23) (319, 4.24)
        };
    
    
        \node[font=\bfseries, anchor=south west] at (axis cs: 200, 6) {JiT-B/16};

    \nextgroupplot[
        ylabel={Guided FID $\downarrow$},
        xmode=log,
        ymode=log,
        xmin=80, xmax=340,
        xtick={50, 100, 150, 200, 250, 300},
        xticklabels={50, 100, 150, 200, 250, 300},
        ymin=2.5, ymax=5,
        ytick={3, 4, 5, 6, 7, 8, 9, 10},
        log ticks with fixed point,
    ]
        \addplot[color=pastelgreen, mark=*] coordinates {
            (59, 8.78) (79, 5.29) (99, 4.55) (119, 3.89) (139, 3.47) (159, 3.23) (179, 3.06) (199, 2.95) (219, 2.84) (239, 2.80) (259, 2.73) (279, 2.66) (299, 2.71) (319, 2.64)
        };
    
        \addplot[color=pastelblue, mark=*] coordinates {
            (59, 5.98) (79, 4.20) (99, 3.67) (119, 3.45) (139, 3.29) (159, 3.09) (179, 2.99) (199, 2.93) (219, 2.84) (239, 2.80) (259, 2.78) (279, 2.70) (299, 2.67) (319, 2.61)
        };
    
    
        \node[font=\bfseries, anchor=south west] at (axis cs: 200, 3.8) {JiT-L/16};

    \nextgroupplot[
        ylabel={Guided FID $\downarrow$},
        xmode=log,
        ymode=log,
        xmin=80, xmax=340,
        xtick={50, 100, 150, 200, 250, 300},
        xticklabels={50, 100, 150, 200, 250, 300},
        ymin=2.4, ymax=4,
        ytick={3, 4, 5, 6, 7, 8, 10, 20, 50, 100, 200},
        log ticks with fixed point,
    ]
        \addplot[color=pastelgreen, mark=*] coordinates {
            (80, 5.22) (100, 4.13) (120, 3.56) (140, 3.32) (160, 3.03)
            (200, 2.88) (220, 2.72) (240, 2.64) (260, 2.56) (280, 2.51)
            (300, 2.54) (320, 2.46)
        };
    
        \addplot[color=pastelblue, mark=*] coordinates {
            (59, 5.86) (79, 4.21) (99, 3.62) (119, 3.27) (139, 3.03) (159, 2.89) (179, 2.81) (199, 2.70) (219, 2.61) (239, 2.59) (259, 2.55) (279, 2.53) (299, 2.46) (319, 2.45)
        };
    
        \node[font=\bfseries, anchor=south west] at (axis cs: 200, 3.3) {JiT-XL/16};
        \end{groupplot}

    \node[
        above=0.3cm of group c2r1.north,
        anchor=south,
        inner sep=3pt,
        draw=gray!50,
        rounded corners=2pt,
    ] (legend) {
        \begin{tikzpicture}
            \begin{axis}[
                hide axis,
                xmin=0, xmax=1, ymin=0, ymax=1,
                legend columns=3,
                legend style={
                    draw=none,
                    fill=none,
                    /tikz/every even column/.append style={column sep=1.5cm},
                    font=\small,
                },
                legend entries={\vloss, \floss}, 
            ]
                \addlegendimage{color=pastelgreen, mark=*, line width=2.5pt, mark size=3.5pt, mark options={fill=white, thick}}
                \addlegendimage{color=pastelblue, mark=*, line width=2.5pt, mark size=3.5pt, mark options={fill=white, thick}}
            \end{axis}
        \end{tikzpicture}
    };

\end{tikzpicture}
    }
    \caption{\textbf{Convergence of frequency- and pixel-space losses.} The frequency-domain loss dominates early in training, while the spatial-domain loss gradually catches up in later stages.}
    \label{fig:freq_training_speed}
\end{figure}

We combine the frequency-domain component with the standard pixel component \vloss into a single  joint objective:

\begin{equation}
    \mathcal{L} = w_f \cdot \mathcal{L}_{f} + w_v \cdot \mathcal{L}_{v}
    \label{eq:vfloss}
\end{equation}

where $w_f$ and $w_v$ are scalars that control the relative contribution of each domain throughout training. Empirically, we observe in Figure~\ref{fig:freq_training_speed} that \floss dominates early in training, converging significantly faster than \vloss. However, \vloss surpasses \floss later in training. It suggests that the two components are not equally useful at all stages: frequency supervision is the binding constraint early on, while pixel supervision becomes more effective as training progresses. We set $w_f = \lambda(s)$ and $w_v = 1 - \lambda(s)$ where $\lambda(s)$ follows a sigmoid schedule decaying from 1 to 0, centered at the observed crossover point $s^\star$. The design choices for the combination schedule are ablated in Section~\ref{subsec:abla}. We call the resulting scheduled loss \vfloss.

\subsection{Empirical Validation}

\begin{figure}[t]
    \centering
    \resizebox{\textwidth}{!}{
        \input{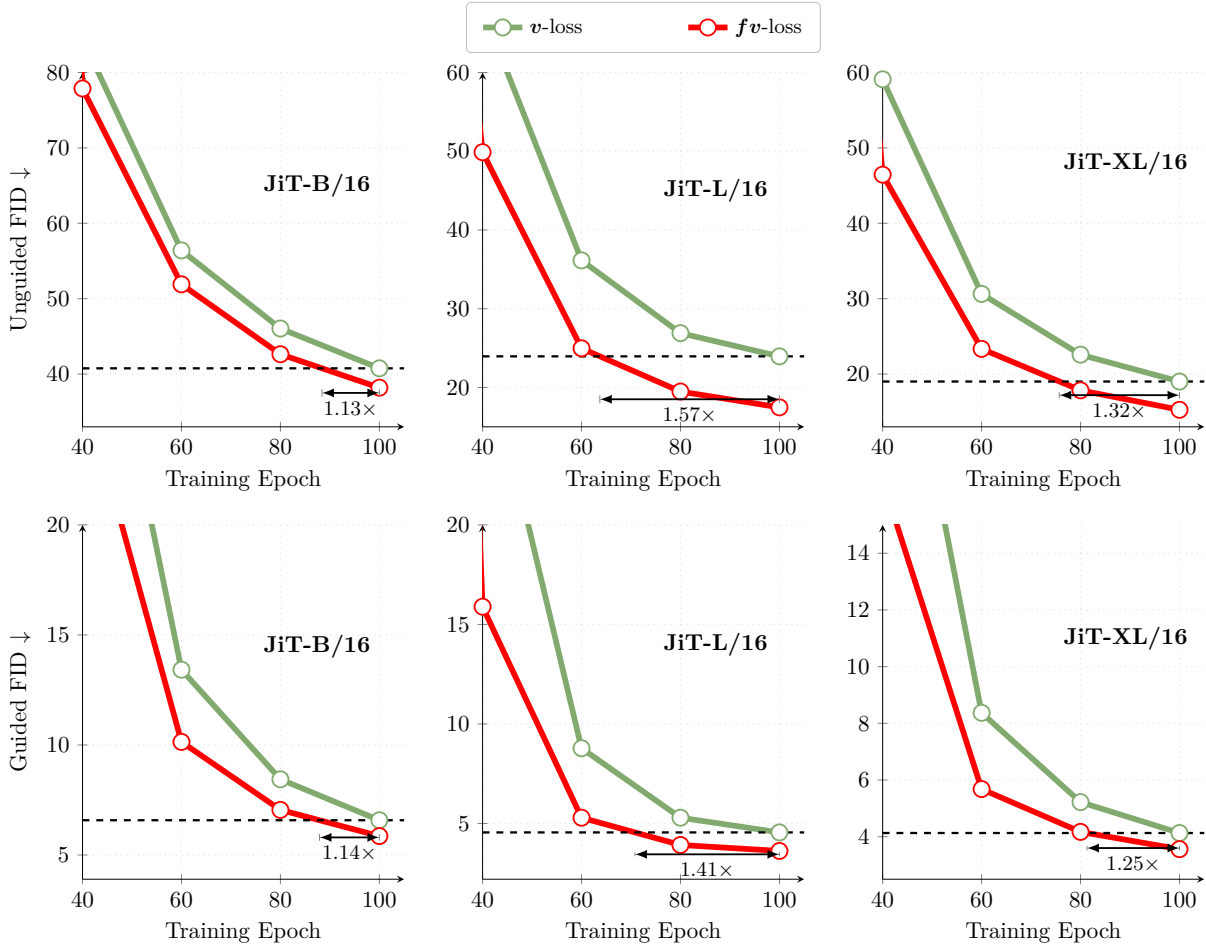}
    }
    \caption{\textbf{Convergence speed comparison} between \vloss and \vfloss for different model sizes. Top row: unguided FID. Bottom row: guided FID. \vfloss consistently outperforms \vloss leading to significant speed-ups.}
    \label{fig:training_speed}
\end{figure}

\paragraph{Convergence speedup} Figure~\ref{fig:training_speed} reports FID across training epochs for JiT-B/16, JiT-L/16, and JiT-XL/16, in both unguided and guided settings. The joint \vfloss objective outperforms the baseline \vloss across all model sizes from the earliest evaluations. The gap is largest in the early epochs and narrows as training progresses, consistent with our observation that frequency supervision is most valuable early in training.

\begin{figure}
    \centering
    \resizebox{\textwidth}{!}{
    \input{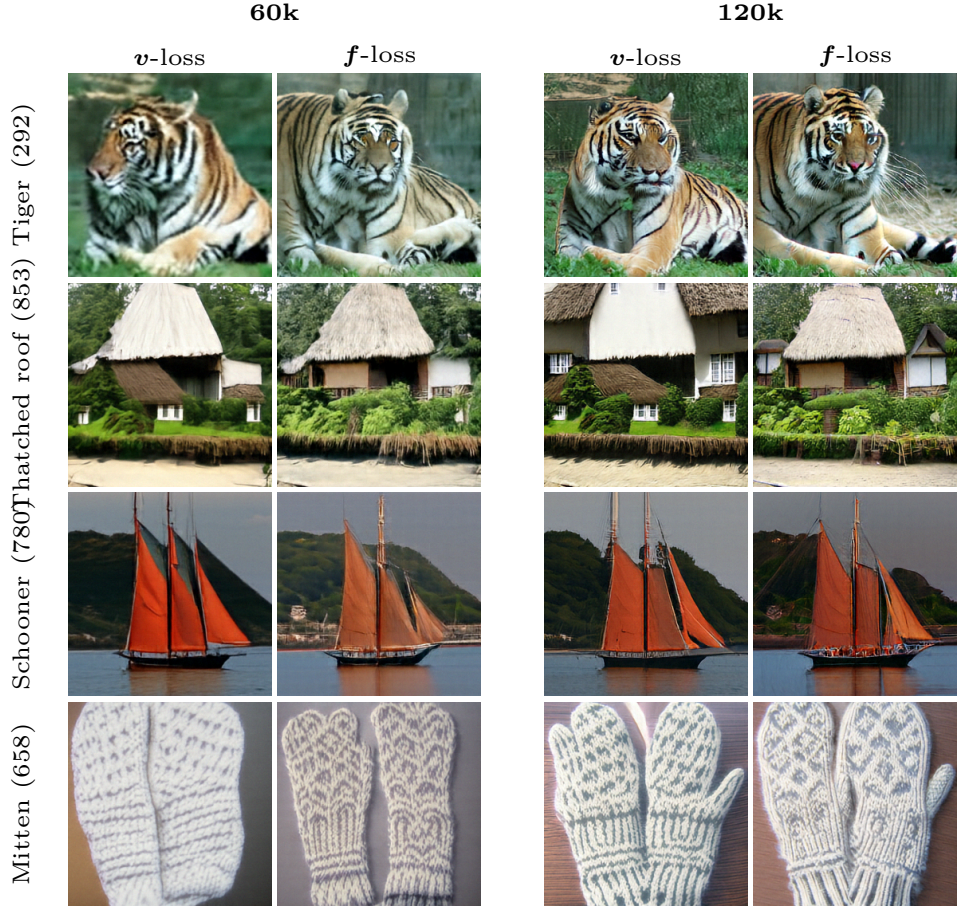}
    }
    \caption{\textbf{Generated images depending on the loss at early training stages.} (left) \floss images have finer textures (\textit{e.g.}, the tiger's stripes, the hay on the thatched roof) much earlier than \vloss that seems to focus more on sharp object boundaries (\textit{e.g.}, well defined sails of the boat) at such early stages. (right) These finer textures starts to appear with \vloss later during training, which may explain its slower start in FID. Please refer to Figure \ref{fig:extended_evolution} for comparison with finer number of steps.}
    \label{fig:evolution}
\end{figure}

\paragraph{Qualitative evidence} Figure~\ref{fig:evolution} compares images generated by \vloss and \floss at two stages of training: very early (left block) and early (right block). The difference is striking at very early training stages. \floss produces images with significantly finer textures well before \vloss: the hay on the thatched roof is detailed, and the knitting pattern on the mittens is already visible. \vloss at the same stage appears to prioritize sharp object boundaries over surface texture. Later in training, \vloss recovers many of these textures, though they remain less defined than those produced by \floss. 

\paragraph{Frequency-Domain Error Analysis} Figure~\ref{fig:wallclock_frequency} (right) measures the average magnitude error between generated and real images in the low- and high-frequency bands of the Fourier spectrum. In the earliest stages of training (50k-75k steps), \floss achieves a lower error than \vloss in both bands, confirming that it corrects the spectral bias faster. As training progresses, \floss plateaus and is eventually overtaken by \vloss. \vfloss combines the fast early spectral alignment of \floss with the late-stage refinement of \vloss, reaching the lowest error in both bands by the end of training.

$f$-loss forces the model to escape this low-frequency bias by actively equalizing the learning signal across all frequency bands. But it is insensitive to phase, which carries the precise spatial location of edges and textures. \vloss directly supervises pixel values and is ultimately better suited for enforcing strict local spatial coherence and phase alignment (e.g., locking sharp edges into exact pixel locations). We hypothesize that this fundamental difference explains why the f-loss eventually plateaus when used alone, making the transition back to the spatial-domain v-loss necessary for final refinement.

\section{Experiments}
 \label{sec:exp}

In this section, we discuss the effectiveness of training an image generative model using our proposed \vfloss. We detail our experimental setup and our evaluation metric to support our experiments. Based on these, we see how our proposed method performs against state-of-the-art models in Section~\ref{sec:sota_comp}. Following that, we show empirical analysis for specific components of our proposed \vfloss and compare it against the standard \vloss to study the convergence dynamics in Section \ref{subsec:abla}. 

\paragraph{Experimental setup.} For fair and reproducible evaluations, we train our models using the ImageNet class conditional setup~\cite{ILSVRC15}; at $256\times 256$ and $512\times 512$ resolutions. We primarily focus on pixel-space flow matching framework following the \texttt{JiT} architecture~\cite{jit} with $x$-pred and a reformulated loss. Unless we state otherwise, models are trained using identical training schedules and compute budgets (batch size: 1024; lr: constant at 2e-4 after 5 epochs of warmup; \smash{optim}: AdamW~\cite{loshchilov2017decoupled} with $\beta_1$, $\beta_2$: 0.9, 0.95), allowing us to isolate the effect of the proposed loss. For evaluation, we generate $50$K samples using 50 ODE steps of Heun sampler~\cite{heun1900neue} and compute all metrics following the standard protocols used in recent generative modeling benchmarks~\cite{dhariwal2021diffusion}. We refer readers to Table \ref{tab:config} for more detailed information about experimental hyperparameters.

\paragraph{Evaluation Metrics.} To evaluate the effectiveness of our proposed method, we primarily use the standard \textbf{Fréchet Inception Distance} (\texttt{FID})~\cite{heusel2017gans} to quantify the quality of the generated samples. \texttt{FID} measures distance between the \texttt{InceptionV3}~\cite{szegedy2016rethinking} feature distributions of generated and real images, providing a standard measure of overall generation quality. Additionally, for our ablations we also evaluate using \textbf{Fréchet Dino Distance} (\texttt{FDD})~\cite{stein2023exposing}, which calculates the Fréchet distance but based on \texttt{DinoV2}~\cite{oquab2024dinov} features. Finally, following prior works~\cite{dhariwal2021diffusion, jit}, we also evaluate using \textbf{Inception Score} (\texttt{IS})~\cite{salimans2016improved} to show that our proposed method does pick up on class-specific information.

\subsection{State-of-the-art comparison}
\label{sec:sota_comp}

We compare our proposed \vfloss against the state-of-the-art pixel flow models in Table \ref{tab:sota}. Our proposed \vfloss improves upon the standard \texttt{JiT} framework w.r.t. the same model architecture (\texttt{XL/16}) and training settings (2.13 vs 2.21 after 750K steps). Following recent methods~\cite{deco,chen2025pixelflow} we also apply \texttt{REPA} alignment loss. Interestingly, with our proposed method we match \texttt{DeCo} performance in less than half the training steps. Finally, when adding perceptual losses similar to \texttt{PixelGen}, our proposed method improves on the current state-of-the-art performance and achieves the same FID of 1.83 in only 500k steps (compared to 800k for PixelGen) while significantly improving on IS, without any additional architectural changes.

\begin{table}
    \centering
    \begin{tabular}{l ccc c cc}
        \toprule
        \textbf{Method} & \textbf{Model size} & \textbf{REPA} & \textbf{Perceptual} & \textbf{Training steps} & \textbf{FID}{\color{gray!80}$\downarrow$} & \textbf{IS} {\color{gray!80}$\uparrow$} \\
        \midrule
        {\color{gray!80}\texttt{JiT-H/16}}~\cite{jit} & {\color{gray!80}H/16} & {\color{gray!80}\xmark} & {\color{gray!80}\xmark} & {\color{gray!80}750k} & {\color{gray!80}1.86} & {\color{gray!80}303.4}  \\ 
        {\color{gray!80}\texttt{JiT-G/16}}~\cite{jit} & {\color{gray!80}G/16} & {\color{gray!80}\xmark} & {\color{gray!80}\xmark} & {\color{gray!80}750k} & {\color{gray!80}1.82} & {\color{gray!80}292.6} \\
        \addlinespace[2pt]
        \cmidrule(lr){1-7}
        \addlinespace[2pt]
        \texttt{JiT-XL/16}~\cite{jit} & XL/16 & \xmark & \xmark & 750k & 2.21 & \textbf{297.4}  \\ 
        \rowcolor{Periwinkle!10}
        \texttt{Ours} & XL/16 & \xmark & \xmark & 750k & \textbf{2.13} & 290.3 \\
        \addlinespace[2pt]
        \cmidrule(lr){1-7}
        \addlinespace[2pt]
        \texttt{EPG}~\cite{lei2026there} {\scriptsize ICLR, 2026} & XL/16 & \cmark &   \xmark    & 1M    & 2.04 & 283.2 \\
        \texttt{PixelFlow}~\cite{chen2025pixelflow} {\scriptsize 2025} & XL/4  & \cmark &  \xmark      & 1.6M  & 1.98 & 282.1 \\
        \texttt{DiP}~\cite{chen2025dip} {\scriptsize CVPR, 2026}    & XL/16 & \cmark &   \xmark     & 1.6M  & 1.98 & 282.9 \\
        \texttt{PixNerd}~\cite{wang2026pixnerd} {\scriptsize ICLR, 2026} & XL/16 & \cmark &   \xmark     & 1.6M  & 1.95 & 298.0 \\
        \texttt{DeCo}~\cite{deco} {\scriptsize CVPR, 2026}    & XL/16 & \cmark &  \xmark      & 1.6M  & 1.90 & \textbf{303.0} \\
        \rowcolor{Periwinkle!10}
        \texttt{Ours}     & XL/16 & \cmark &   \xmark     & 750k  & \textbf{1.87} & 301.0 \\
        \addlinespace[2pt]
        \cmidrule(lr){1-7}
        \addlinespace[2pt]
        \texttt{PixelGen}~\cite{pixelgen} {\scriptsize 2026} & XL/16 & \cmark & \cmark & 800k  & \textbf{1.83} & 293.6 \\
        \rowcolor{Periwinkle!10}
        \texttt{Ours}     & XL/16 & \cmark & \cmark & 500k  & \textbf{1.83} & \textbf{323.4} \\
        \bottomrule
    \end{tabular}
    \vspace{0.1cm}
    \caption{\textbf{Comparison with state-of-the-art pixel-space methods.} Our proposed loss is able to improve over all categories of models (standard, using alignment methods like REPA, using perceptual losses), either by achieving strictly better FIDs or by reaching the same FID faster.}
    \label{tab:sota}
\end{table}

\begin{table}[]
    \centering
    \begin{subtable}[t]{0.48\linewidth}
        \centering
        \resizebox{\linewidth}{!}{%
        \begin{tabular}{@{}l c c cc cc}
            \toprule
            & \textbf{Epochs} & \textbf{Loss} & \multicolumn{2}{c}{\textbf{FID} {\color{gray!80}$\downarrow$}} & \textbf{FDD} {\color{gray!80}$\downarrow$}& \textbf{IS} {\color{gray!80}$\uparrow$} \\
            \cmidrule(lr{0.5em}){4-5} \cmidrule(lr{0.5em}){6-6} \cmidrule(lr{0.5em}){7-7}
            & & & \footnotesize{w/o cfg} & \footnotesize{cfg} & \footnotesize{cfg} & \footnotesize{cfg} \\
            \midrule
            \multirow{9}{*}{\rotatebox[origin=c]{90}{\texttt{JiT-B/16}}} & \multirow{3}{*}{80}  & \vlosstablecolor  & 46.02 & 8.44 & 369.3 & 162.8 \\
            & & \flosstablecolor  & \textbf{42.34} & \textbf{6.82} & \textbf{312.0} & \textbf{186.6} \\
            & & \vflosstablecolor & 42.63 & 7.05 & 314.0 & 185.2 \\
            \addlinespace[2pt] 
            \cmidrule(lr){2-2}\cmidrule(lr){3-3}\cmidrule(lr){4-5}\cmidrule(lr){6-6}\cmidrule(lr){7-7}
            \addlinespace[2pt]
            & \multirow{3}{*}{200} & \vlosstablecolor  & 31.23 & 4.52 & 219.8 & 252.1 \\
            & & \flosstablecolor  & \textbf{30.02} & 4.54 & \textbf{194.8} & 254.3 \\
            & & \vflosstablecolor & 30.17 & \textbf{4.20} & 202.3 & \textbf{258.9} \\
            \addlinespace[2pt]
            \cmidrule(lr){2-2}\cmidrule(lr){3-3}\cmidrule(lr){4-5}\cmidrule(lr){6-6}\cmidrule(lr){7-7}
            \addlinespace[2pt]
            & \multirow{3}{*}{320} & \vlosstablecolor  & 28.70 & 4.11 & 192.2 & 273.9 \\
            & & \flosstablecolor  & 27.51 & 4.23 & \textbf{170.4} & 275.2 \\
            & & \vflosstablecolor & \textbf{26.86} & \textbf{3.95} & 178.8 & \textbf{282.6} \\
            \addlinespace[2pt]
            \cmidrule(lr{0.5em}){1-7}
            \addlinespace[2pt]
            \multirow{9}{*}{\rotatebox[origin=c]{90}{\texttt{JiT-L/16}}} & \multirow{3}{*}{80}  & \vlosstablecolor  & 26.90 & 5.40 & 264.2 & 200.0 \\
            & & \flosstablecolor  & 20.24 & 4.34 & 185.8 & 237.3 \\
            & & \vflosstablecolor & \textbf{19.47} & \textbf{3.93} & \textbf{178.6} & \textbf{247.9} \\
            \addlinespace[2pt]
            \cmidrule(lr){2-2}\cmidrule(lr){3-3}\cmidrule(lr){4-5}\cmidrule(lr){6-6}\cmidrule(lr){7-7}
            \addlinespace[2pt]
            & \multirow{3}{*}{200} & \vlosstablecolor  & 16.54 & 2.98 & 155.4 & 274.3 \\
            & & \flosstablecolor  & \textbf{14.14} & 2.92 & 125.1 & 283.3 \\
            & & \vflosstablecolor & 14.79 & \textbf{2.82} & \textbf{122.2} & \textbf{291.6} \\
            \addlinespace[2pt]
            \cmidrule(lr){2-2}\cmidrule(lr){3-3}\cmidrule(lr){4-5}\cmidrule(lr){6-6}\cmidrule(lr){7-7}
            \addlinespace[2pt]
            & \multirow{3}{*}{320} & \vlosstablecolor  & 14.46 & 2.63 & 133.7 & 292.7 \\
            & & \flosstablecolor  & 13.26 & 2.58 & 115.2 & 287.5 \\
            & & \vflosstablecolor & \textbf{12.93} & \textbf{2.55} & \textbf{113.7} & \textbf{303.4} \\
            \bottomrule
        \end{tabular}%
}
        \caption{Resolution $256\times 256$}
        \label{tab:baselinefocal_256}
    \end{subtable}
    \hfill
    \begin{subtable}[t]{0.48\linewidth}
        \centering
        \resizebox{\linewidth}{!}{%
        \begin{tabular}{@{}l c c cc cc}
            \toprule
            & \textbf{Epochs} & \textbf{Loss} & \multicolumn{2}{c}{\textbf{FID} {\color{gray!80}$\downarrow$}} & \textbf{FDD}{\color{gray!80}$\downarrow$}  & \textbf{IS} {\color{gray!80}$\uparrow$} \\
            \cmidrule(lr{0.5em}){4-5} \cmidrule(lr{0.5em}){6-6} \cmidrule(lr{0.5em}){7-7}
            & & & \footnotesize{w/o cfg} & \footnotesize{cfg} & \footnotesize{cfg} & \footnotesize{cfg} \\
            \midrule
            \multirow{9}{*}{\rotatebox[origin=c]{90}{\texttt{JiT-B/32}}} & \multirow{3}{*}{80}  & \vlosstablecolor  & 52.72 & 10.40 & 833.8 & 153.7 \\
            & & \flosstablecolor  & 50.54 & 9.48  & 792.3 & 159.0 \\
            & & \vflosstablecolor &\textbf{ 45.90} &\textbf{ 8.33}  & \textbf{775.1} & \textbf{175.2} \\
            \addlinespace[2pt]
            \cmidrule(lr){2-2}\cmidrule(lr){3-3}\cmidrule(lr){4-5}\cmidrule(lr){6-6}\cmidrule(lr){7-7}
            \addlinespace[2pt]
            & \multirow{3}{*}{200} & \vlosstablecolor  & 35.20 & 5.18 & 676.4 & 244.5 \\
            & & \flosstablecolor  & 36.54 & 5.19 & 664.1 & 231.9 \\
            & & \vflosstablecolor & \textbf{33.12} & \textbf{4.74} & \textbf{653.7} & \textbf{254.3} \\
            \addlinespace[2pt]
            \cmidrule(lr){2-2}\cmidrule(lr){3-3}\cmidrule(lr){4-5}\cmidrule(lr){6-6}\cmidrule(lr){7-7}
            \addlinespace[2pt]
            & \multirow{3}{*}{320} & \vlosstablecolor  & 31.94 & 4.63 & 642.8 & 270.1 \\
            & & \flosstablecolor  & 32.93 & 4.64 & 634.5 & 254.9 \\
            & & \vflosstablecolor & \textbf{29.12} & \textbf{4.47} & \textbf{626.9} & \textbf{281.9} \\
            \addlinespace[2pt]
            \cmidrule(lr{0.5em}){1-7}
            \addlinespace[2pt]
            \multirow{9}{*}{\rotatebox[origin=c]{90}{\texttt{JiT-L/32}}} & \multirow{3}{*}{80}  & \vlosstablecolor  & 29.04 & 6.41 & 746.4 & 228.2 \\
            & & \flosstablecolor  & 22.91 & 4.65 & 689.5 & \textbf{237.0} \\
            & & \vflosstablecolor & \textbf{21.64} & \textbf{4.58} & \textbf{681.9} & 196.9 \\
            \addlinespace[2pt]
            \cmidrule(lr){2-2}\cmidrule(lr){3-3}\cmidrule(lr){4-5}\cmidrule(lr){6-6}\cmidrule(lr){7-7}
            \addlinespace[2pt]
            & \multirow{3}{*}{200} & \vlosstablecolor  & 17.80 & 3.22 & 650.6 & \textbf{288.8} \\
            & & \flosstablecolor  & 16.71 & 3.26 & 633.3 & 279.1 \\
            & & \vflosstablecolor & \textbf{15.84} & \textbf{2.98} & \textbf{631.0} & 284.5 \\
            \addlinespace[2pt]
            \cmidrule(lr){2-2}\cmidrule(lr){3-3}\cmidrule(lr){4-5}\cmidrule(lr){6-6}\cmidrule(lr){7-7}
            \addlinespace[2pt]
            & \multirow{3}{*}{320} & \vlosstablecolor  & 15.85 & 2.90 & 639.8 & 294.6 \\
            & & \flosstablecolor  & 15.28 & 3.04 & \textbf{619.5} & 282.4 \\
            & & \vflosstablecolor & \textbf{14.30} & \textbf{2.73} & 620.2 & \textbf{312.1} \\
            \bottomrule
        \end{tabular}%
        }
        \caption{Resolution $512\times 512$}
        \label{tab:baselinefocal_512}
    \end{subtable}
    \vspace{0.2cm}
    \caption{\textbf{Comparison between \vlosstablecolor, \flosstablecolor and \vflosstablecolor} at resolutions (a) $256\times256$ and (b) $512\times512$. \vflosstablecolor combines the best of \vlosstablecolor and \flosstablecolor, namely speed of convergence at early stages and final performances.}
    \label{tab:baselinefocal_combined}
\end{table}

\subsection{Ablation Study}
\label{subsec:abla}

We begin by first ablating our proposed \vfloss (given in Equation \ref{eq:vfloss}). In the $256^2$ resolution setup for a \texttt{B} size model (see Table \ref{tab:baselinefocal_256}), we see that using simple \floss over \vloss is highly beneficial early in the training for both unguided ($42.34$ vs $46.02$) and guided ($6.82$ vs $8.44$) settings. As training progresses, the \floss is still better than \vloss w.r.t unguided setup ($27.51$ vs $28.70$). A similar trend can also be seen for bigger \texttt{L} size models too, where \floss is better than the \vloss early in the training for unguided settings ($20.24$ vs $26.90$ at 80 epochs) and remains on par at the end ($13.26$ vs $14.46$ at 320 epochs). However, for the guided setup, the \vloss becomes better than the \floss at later training stage, irrespective of the model size. We argue that this is because, as training progresses, it is more important for the model to learn from the pixels to improve upon the high-frequency details. Interestingly, with this in mind our proposed \vfloss setup with a switch from frequency to spatial domain shows improvement both in the early stages of training as well as in the later stages for both unguided and guided setups.

A similar trend can also be observed for $512^2$ resolution experiments (see Table \ref{tab:baselinefocal_512}). In the early stages of training \floss is always better than \vloss irrespective of model sizes (for \texttt{B}: 9.48 vs 10.40; for \texttt{L}: 4.65 vs 6.41 at 80 epochs). However, with training progression, a switch using \vfloss is much better compared to \vloss and \floss alone. 

This provides empirical evidence that, indeed, beginning with a much more balanced frequency-based objective is beneficial for better pixel space models. However, at the end, where there is much more need for fine-grained details, reverting back to a spatial loss is highly beneficial.

\paragraph{Adaptative weighting} To understand the switch in \floss as given in Equation \ref{eq:vfloss}, we also ablate the effect of $w_f$ and $w_v$. As reported in Table \ref{tab:weights_ablatiion}, clearly having a sigmoid weight performs the best throughout training, w.r.t. \texttt{FID}. Interestingly, a higher weight on \vloss also helps but is not as good as the proposed sigmoid switch.
Crucially, a reversed sigmoid that emphasizes mostly pixel loss at the beginning and frequencies loss at later stage has among the worst performances.
This clearly indicates that a stronger pixel flow model requires some spectral balancing in the early training stage to set the stage for spatial high-detail learning.

\begin{table}[]
    \centering

\begin{tabular}{cc cccc}
    \toprule
    \boldmath{$w_f$} & \boldmath{$w_v$} & \multicolumn{3}{c}{\textbf{FID} {\color{gray!80}$\downarrow$}} & \textbf{IS} {\color{gray!80}$\uparrow$} \\
    \cmidrule(lr){3-5} \cmidrule(lr){6-6}
    & & \footnotesize{Epoch 80} & \footnotesize{Epoch 200} & \footnotesize{Epoch 320} & \\
    \cmidrule(lr){1-6}
    1 & 0 & 6.82 & 4.54 & 4.23 & 275.18 \\
    0 & 1 & 8.44 & 4.52 & 4.11 & 273.85 \\
    \addlinespace[2pt]
    \cmidrule(lr){1-1}\cmidrule(lr){2-2}\cmidrule(lr){3-5}\cmidrule(lr){6-6}
    \addlinespace[2pt]
    \multirow{5}{*}{1} & 0.1 & 7.24 & 4.45 & 4.16 & 269.16 \\
     & 0.5 & 6.85 & 4.49 & 4.31 & 279.18 \\
     & 1   & 7.04 & 4.33 & 4.05 & 274.85 \\
     & 5   & 7.05 & 4.51 & 4.22 & 282.47 \\
     & 10  & 7.14 & 4.25 & 4.02 & \textbf{285.88} \\
    \addlinespace[2pt]
    \cmidrule(lr){1-1}\cmidrule(lr){2-2}\cmidrule(lr){3-5}\cmidrule(lr){6-6}
    \addlinespace[2pt]
    $1-\lambda(s)$ & $\lambda(s)$ & 8.52 & 4.48 & 4.30 & 265.9 \\
    \rowcolor{Periwinkle!10}
    $\lambda(s)$ & $1-\lambda(s)$ & 7.05 & 4.20 & \textbf{3.95} & 282.58 \\
    \bottomrule
\end{tabular}
    \vspace{0.1cm}
    \caption{Ablation on the weights $w_f$ and $w_v$ of \vfloss}
    \label{tab:weights_ablatiion}
\end{table}

\paragraph{Generalization across architectures.} To test whether the benefits of \vfloss generalize beyond JiT, we train PixelDiT~\cite{pixeldit} with $v$-loss and \vfloss following the original PixelDiT setup. Table~\ref{tab:pixeldit} shows that \vfloss reaches the same FID roughly $2\times$ faster than $v$-loss. Moreover, PixelDiT is trained with the REPA alignment loss. REPA pushed the features toward rich high-frequencies DINO features. \vfloss is beneficial even when used alongside REPA.

\begin{table}[h]
    \centering
    \small
    \begin{tabular}{@{}l l ccc}
        \toprule
        & \textbf{Loss} & \textbf{Step 200k} & \textbf{Step 400k} & \textbf{Step 600k} \\
        \midrule
        \multirow{3}{*}{\rotatebox[origin=c]{90}{\tiny \texttt{PixelDiT}}} & \vlosstablecolor (reported from \cite{pixeldit}) & 15 & 12 & 8 \\
        & \vlosstablecolor (reproduced) & 15.09 & 10.60 & 10.25 \\
        & \vflosstablecolor & \textbf{10.18} & \textbf{6.88} & \textbf{5.92} \\
        \bottomrule
    \end{tabular}
    \vspace{0.1cm}
    \caption{\textbf{FID Comparison of \vlosstablecolor and \vflosstablecolor using PixelDiT~\cite{pixeldit} architecture.}}
    \label{tab:pixeldit}
\end{table}



\begin{figure}[]
    \centering
    \resizebox{\textwidth}{!}{


\definecolor{pastelgreen}{RGB}{119, 180, 122}
\definecolor{pastelblue}{RGB}{123, 163, 206}

\begin{tikzpicture}

\pgfmathsetmacro{\leftW}{4.6}      
\pgfmathsetmacro{\rightW}{4.2}     
\pgfmathsetmacro{\rightH}{2.5}     
\pgfmathsetmacro{\rowgap}{0.3}     
\pgfmathsetmacro{\colgap}{1.5}     
\pgfmathsetmacro{\leftH}{2*\rightH+\rowgap}

\pgfmathsetmacro{\insetW}{2.5}     
\pgfmathsetmacro{\insetH}{3.2}     
\pgfmathsetmacro{\insetX}{-0.05}   
\pgfmathsetmacro{\insetY}{-0.15}   

\pgfmathsetmacro{\legendgap}{0.15}  
\pgfmathsetmacro{\legendsep}{0.7}  

\pgfmathsetmacro{\lineW}{1.8}      
\pgfmathsetmacro{\markS}{2.4}      
\pgfmathsetmacro{\inlineW}{1.1}    
\pgfmathsetmacro{\inmarkS}{1.4}    

\pgfmathsetmacro{\titleX}{0.98}
\pgfmathsetmacro{\titleY}{0.89}
\newcommand{\titlefont}{\footnotesize}
\newcommand{\inplottitle}[1]{%
    \node[anchor=north east, font=\titlefont, inner sep=1.5pt]
        at (rel axis cs:\titleX,\titleY) {#1};%
}

\pgfplotsset{
    scale only axis,
    axis lines=left,
    grid=major,
    grid style={dotted, gray!30},
    label style={font=\footnotesize},
    tick label style={font=\footnotesize},
    xlabel near ticks,
    ylabel near ticks,
    every axis plot/.append style={
        line width=\lineW pt,
        mark=*,
        mark size=\markS pt,
        mark options={fill=white, thick}
    },
}

\begin{axis}[
    name=mainA,
    width=\leftW cm, height=\leftH cm,
    xlabel={Wall-clock training time (min)},
    ylabel={Guided FID},
    xtick={100,200,300,400},
    ymin=3.8, ymax=10.5,
]

    \draw[dashed, gray, fill=gray, fill opacity=0.08]
        (axis cs:250,3.85) rectangle (axis cs:400,5.0);

    \addplot[color=pastelgreen] coordinates {
        (100, 8.44) (200, 5.26) (300, 4.40) (400, 4.14)
    };
    \addplot[color=pastelblue] coordinates {
        (100, 10.24) (200, 5.10) (300, 4.60) (400, 4.57)
    };
    \addplot[color=red] coordinates {
        (100, 10.14) (200, 5.24) (300, 4.32) (400, 3.99)
    };
\end{axis}

\begin{axis}[
    name=insetA,
    at={($(mainA.north east)+(\insetX cm,\insetY cm)$)},
    anchor=north east,
    width=\insetW cm, height=\insetH cm,
    axis lines=box,
    axis background/.style={fill=white},
    axis line style={gray},
    xmin=250, xmax=440,
    ymin=3.85, ymax=4.85,
    xtick={300,400},
    ytick={4.0,4.5},
    tick label style={font=\tiny},
    tick style={gray},
    grid style={dotted, gray!25},
    every axis plot/.append style={
        line width=\inlineW pt, mark=*, mark size=\inmarkS pt,
        mark options={fill=white, thick}
    },
]
    \addplot[color=pastelgreen] coordinates {(200,5.26) (300,4.40) (400,4.14)};
    \addplot[color=pastelblue]  coordinates {(200,5.10) (300,4.60) (400,4.57)};
    \addplot[color=red]         coordinates {(200,5.24) (300,4.32) (400,3.99)};

    \node[anchor=south east, font=\tiny, color=pastelgreen] at (axis cs:430,4.16) {370k};
    \node[anchor=north east, font=\tiny, color=pastelblue]  at (axis cs:430,4.55) {353k};
    \node[anchor=south east, font=\tiny, color=red]         at (axis cs:398,3.9) {318k};
\end{axis}

\begin{axis}[
    name=freqLow,
    at={($(mainA.north east)+(\colgap cm,0)$)},
    anchor=north west,
    width=\rightW cm, height=\rightH cm,
    ylabel={Avg.\ magn.\ error},
    xtick={100,200,300,400},
    xticklabels={},
    ymin=3, ymax=16.5,
]
    \inplottitle{Low frequencies}

    \addplot[color=pastelgreen] coordinates {
        (50, 14.00) (75, 8.87) (100, 6.24) (150, 5.74)
        (200, 4.61) (250, 4.67) (300, 4.51) (400, 4.15)
    };
    \addplot[color=pastelblue] coordinates {
        (50, 12.34) (75, 7.96) (100, 7.07) (150, 6.64)
        (200, 6.26) (250, 6.44) (300, 5.94) (400, 6.03)
    };
    \addplot[color=red] coordinates {
        (50, 12.35) (75, 7.72) (100, 7.03) (150, 6.36)
        (200, 5.35) (250, 4.41) (300, 3.85) (400, 3.79)
    };
\end{axis}

\begin{axis}[
    name=freqHigh,
    at={($(freqLow.south west)+(0,-\rowgap cm)$)},
    anchor=north west,
    width=\rightW cm, height=\rightH cm,
    xlabel={Training step},
    ylabel={Avg.\ magn.\ error},
    xtick={100,200,300,400},
    xticklabels={100k,200k,300k,400k},
    ymin=3.2, ymax=4.8,
]
    \inplottitle{High frequencies}

    \addplot[color=pastelgreen] coordinates {
        (50, 4.51) (75, 4.18) (100, 3.83) (150, 3.69)
        (200, 3.50) (250, 3.48) (300, 3.48) (400, 3.38)
    };
    \addplot[color=pastelblue] coordinates {
        (50, 4.45) (75, 4.06) (100, 3.88) (150, 3.76)
        (200, 3.65) (250, 3.62) (300, 3.53) (400, 3.49)
    };
    \addplot[color=red] coordinates {
        (50, 4.46) (75, 4.07) (100, 3.90) (150, 3.72)
        (200, 3.55) (250, 3.45) (300, 3.36) (400, 3.34)
    };
\end{axis}

\coordinate (legend-anchor) at ($(mainA.north west)!0.5!(freqLow.north east)$);
\node[
    above=\legendgap cm of legend-anchor,
    anchor=south,
    inner sep=1.5pt,
    draw=gray!50,
    rounded corners=2pt,
] (legend) {
    \begin{tikzpicture}
        \begin{axis}[
            hide axis,
            xmin=0, xmax=1, ymin=0, ymax=1,
            legend columns=3,
            legend style={
                draw=none,
                fill=none,
                /tikz/every even column/.append style={column sep=\legendsep cm},
                font=\footnotesize,
            },
            legend entries={$v$-loss, $f$-loss, $fv$-loss},
        ]
            \addlegendimage{color=pastelgreen, mark=*, line width=\lineW pt, mark size=\markS pt, mark options={fill=white, thick}}
            \addlegendimage{color=pastelblue, mark=*, line width=\lineW pt, mark size=\markS pt, mark options={fill=white, thick}}
            \addlegendimage{color=red, mark=*, line width=\lineW pt, mark size=\markS pt, mark options={fill=white, thick}}
        \end{axis}
    \end{tikzpicture}
};

\end{tikzpicture}
    }
    \caption{\textbf{Left:} Guided FID against wall-clock training time; the number of training steps is given for the last point of each curve. \textbf{Right:} Average magnitude error in the low- and high-frequency bands. \vfloss consistently outperforms \vloss, leading to significant speed-ups.}
    \label{fig:wallclock_frequency}
\end{figure}
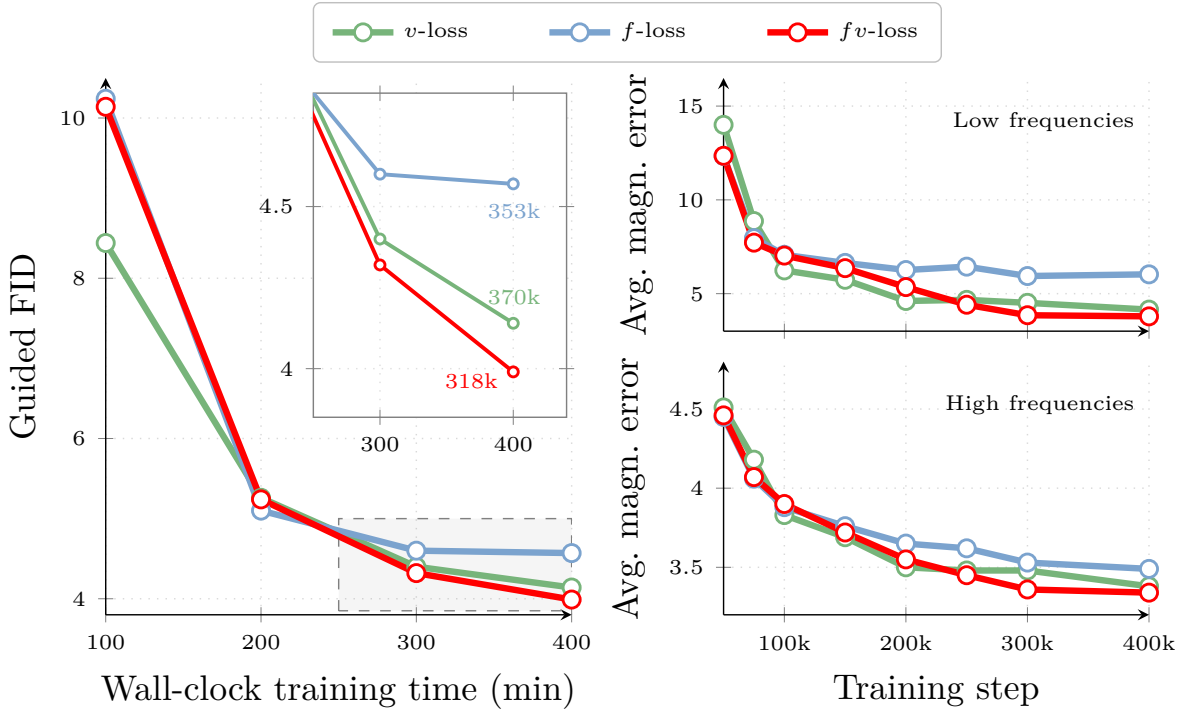

\paragraph{Wall-clock speedup.} Figure~\ref{fig:wallclock_frequency} reports the guided FID reached by each loss at fixed wall-clock budgets, using the training step closest to each time mark. As \floss requires computing a 2D FFT at every training step, it introduces a computational overhead. To verify that this does not negate the reported convergence speed-up, we measure the total wall-clock time for training JiT-B for $400k$ steps. Compared to the \vloss baseline ($432.0$min), \floss incurs a $+4\%$ overhead ($453.1$min). \vfloss exhibits a $+14\%$ overhead ($492.3$min), as it computes and backpropagates both losses at each step. At the 100-minute mark, \vloss is temporarily ahead because \vfloss has completed fewer steps ($90k$ vs. $78k$ steps). However, \vfloss overtakes \vloss by the 200-minute mark. From 300 minutes onward, its convergence speed-up more than compensates for the added computational cost.

\vspace{-0.2cm}

\section{Conclusion}
\label{sec:conclusion}
In this paper, we investigate spectral bias in flow-matching models and show that the pixel-space velocity loss of JiT leads to overestimating low frequencies early during training, at the cost of underestimating higher frequencies. This can be explained by the natural $1/f^2$ decay of the power spectrum of natural images that leads to imbalance in the training objective. This bias towards low frequencies at early training stages seems to be hurting early convergence. As such, we propose a new loss in the frequency domain that aims at mitigating such bias by equalizing frequencies. We experimentally show that such loss obtains better performances at early training stages. We propose to combine the early speed advantage of the frequency loss with the later training performances of the classical pixel-space velocity loss through a schedule. The resulting \vfloss significantly accelerate the convergence of flow-matching training and leads to state-of-the-art pixel-space image generation on class-conditional ImageNet.

\paragraph{Limitations} Whereas the spectral bias of modern pixel space flow-matching models is clearly exposed and a mitigation strategy is successfully proposed, understanding its origin remains an open question. More specifically, future work should investigate why having a frequency objective at early training stages plateaus in performances when used alone, and why switching to pixel space losses performs better at later training stages.
\renewcommand{\thefootnote}{}
\footnotetext{
    \rotatebox{180}{The real images in the teaser are the tiger and the magpie.}
}

\begin{ack}
This work was supported by a Hi!Paris grant, two Hi!Paris chaires for A.A.Efros and V.Kalogeiton, ANR/France 2030 program (ANR-23-IACL-0005) and ANR project sharp ANR-23-
PEIA-0008 in the context of the PEPR IA. It was granted access to the HPC resources of IDRIS under the allocations 2025-A0181016194, 2025-AD011015436, 2026-A0201017545 and 2026-AD011015594R2 made by GENCI. We would like to thank Nicolas Dufour, Julie Mordacq, Robin Courant, Yohann Perron, Adrien Ramanana Rahary and Hovhannes Margaryan for their helpful comments and proofreading.
\end{ack}

\bibliography{bibliography}
\bibliographystyle{plainnat}


\newpage
\appendix

\section{Implementation Details}

Our implementation closely follows the public codebases of JiT \cite{jit}. Table~\ref{tab:config} details the architecture, training, and sampling parameters. All experiments were run on 4 nodes of 4 H100 GPUs. The final XL models were trained on 4 nodes of 4 GB200 GPUs.

We select the switch point $s^\star$ of Equation~\ref{eq:vfloss} using a simple crossing-point rule: we identify where the \floss and \vloss FID curves intersect on a held-out validation set (around 200k steps for the B and L models, and around 350k steps for the XL models), and do not perform a grid search over $s^\star$.

\begin{table}[h!]
    \centering
    \resizebox{1.0\width}{!}{
    \begin{tabular}{l@{\;\;\;}c@{\;\;\;}!{\color{gray!60}\vrule}@{\;\;\;}c@{\;\;\;}!{\color{gray!60}\vrule}@{\;\;}c}
        \toprule
        & \textbf{JiT-B} & \textbf{JiT-L} & \textbf{JiT-XL} \\
        \midrule
        \rowcolor[gray]{0.9}\multicolumn{4}{c}{\textbf{architecture}} \\
        depth & 12 & 24 & 28 \\
        hidden dim & 768 & 1024 & 1152 \\
        heads & 12 & 16 & 16 \\
        patch size & \multicolumn{3}{c}{\texttt{image\_size} / 16} \\
        bottleneck & \multicolumn{3}{c}{128} \\
        in-context class tokens & \multicolumn{3}{c}{32} \\
        in-context start block & 4 & 8 & 8 \\
        \rowcolor[gray]{0.9}\multicolumn{4}{c}{\textbf{training}} \\
        warmup epochs \cite{Goyal2017} & \multicolumn{3}{c}{5} \\
        optimizer & \multicolumn{3}{c}{AdamW \cite{loshchilov2017decoupled}, $\beta_1, \beta_2=0.9, 0.95$} \\
        batch size & \multicolumn{3}{c}{1024} \\
        learning rate & \multicolumn{3}{c}{2e-4} \\
        learning rate schedule & \multicolumn{3}{c}{constant} \\
        weight decay & \multicolumn{3}{c}{0} \\
        ema decay & \multicolumn{3}{c}{0.9999} \\
        time sampler & \multicolumn{3}{c}{$\text{logit}(t){\sim}\mathcal{N}(\mu, \sigma^2)$, $\mu$ = -0.8, $\sigma$ = 0.8} \\
        noise scale & \multicolumn{3}{c}{1.0 $\times$ \texttt{image\_size} / 256} \\
        clip of $(1-t)$ in division & \multicolumn{3}{c}{0.05} \\
        class token drop (for CFG) & \multicolumn{3}{c}{0.1} \\
        \rowcolor[gray]{0.9}\multicolumn{4}{c}{\textbf{sampling}} \\
        ODE solver & \multicolumn{3}{c}{Heun \cite{heun1900neue}} \\
        ODE steps & \multicolumn{3}{c}{50} \\
        time steps & \multicolumn{3}{c}{linear in [0.0, 1.0]} \\
        CFG scale sweep range \cite{ho2022classifier} & \multicolumn{3}{c}{[1.0, 4.0]} \\
        CFG interval \cite{kynkaanniemi2024applying} & \multicolumn{3}{c}{[0.1, 1] (if used)} \\
        \bottomrule
    \end{tabular}
    }
    \vspace{0.1cm}
    \caption{\textbf{Configurations of experiments.}}
    \label{tab:config}
\end{table}

\section{Additional Results}

\subsection{\texorpdfstring{Design choice of $f$-loss and $fv$-loss}{Design choice of f-loss and fv-loss}}
\label{sec:abl_log_ffl}

In this section, we ablate the design components of \floss{} using a B-size model and report the results in Table~\ref{tab:focal}. Note that those ablations have been done with cfg interval, RoPE and in-context tokens.
First, we compare against the original focal frequency loss~\cite{ffl} (first row), which performs significantly worse than our formulation. The key difference between the two formulations is the logarithmic compression of the spectral residual: while the loss of~\cite{ffl} reweights the raw magnitude error, \floss compresses it with $\log(1+e_{u,v})$ before reweighting, which we ground as a continuous generalization of a Laplacian pyramid (Appendix~\ref{app:laplacian}): it assigns comparable structural importance to each octave instead of letting the largest residuals dominate the loss.
Introducing the logarithmic compression of spectral residuals already yields substantial improvements, confirming that the $\log$ transformation (as motivated by the scale hierarchy of Laplacian pyramids) plays a key role in stabilizing the spectral balancing.

We further observe that normalized weighting schemes consistently outperform their unnormalized counterparts. 
Without normalization, the loss magnitude becomes dominated by a subset of frequencies, preventing effective balancing across the spectrum. 
In contrast, normalizing the spectral weights ensures that the relative difficulty of different frequencies is preserved during optimization. 
Combining both components yields the best performance, achieving an FID of $41.78$ (cfg=1.0) and $6.39$ (cfg=3.5).

\begin{table}[b]
    \centering
    \resizebox{\textwidth}{!}{
    \begin{tabular}{l@{\;\;\;\;}!{\color{gray!60}\vrule}@{\;\;}c@{\;\;}!{\color{gray!60}\vrule}@{\;\;}c@{\;\;}!{\color{gray!60}\vrule}@{\;\;}c@{\;\;}!{\color{gray!60}\vrule}@{\;\;}c@{\;\;}!{\color{gray!60}\vrule}@{\;\;}c@{\;\;}!{\color{gray!60}\vrule}@{\;\;}c@{\;\;}!{\color{gray!60}\vrule}@{\;\;}c@{\;\;}!{\color{gray!60}\vrule}@{\;\;}c@{\;\;}}
        \toprule
        \textbf{Loss} & $w_{u, v}$ & $\log(1 + e_{u, v})$ & \multicolumn{2}{c@{\;\;}!{\color{gray!60}\vrule}@{\;\;}}{\textbf{FID} {\color{gray!80}$\downarrow$}} & \textbf{Prec.} {\color{gray!80}$\uparrow$} & \textbf{Rec.} {\color{gray!80}$\uparrow$} & \textbf{Den.} {\color{gray!80}$\uparrow$}& \textbf{Cov.} {\color{gray!80}$\uparrow$} \\
        & & & \footnotesize{cfg=1.0} & \footnotesize{cfg=3.5} & & & & \\
        \midrule
        \texttt{focal}~\cite{ffl} & $\frac{e_k}{\max_{k'} e_{k'}}$ & \xmark & 72.80 & 19.24 & 0.38 & 0.54 & 0.32 & 0.60 \\
        \greyrule
        \multirow{5}{*}{\flosstablecolorttt} & 1 & \xmark & 50.12 & 12.67 & 0.49 & 0.63 & 0.48 & 0.77 \\
        \greycmidrule{2-9}
        & 1 & \cmark & 57.17 & 17.01 & 0.43 & 0.61 & 0.39 & 0.72 \\
        \greycmidrule{2-9}
         & $\frac{\log(1+e_k)}{\max_{k'} \log(1+e_k')}$ & 
         \cmark & 44.40 & 7.42 & 0.55 & 0.65 & 0.61 & 0.83 \\
         \greycmidrule{2-9}
         & \cellcolor{Periwinkle!10}$\frac{e_k}{\max_{k'} e_{k'}}$ & \cellcolor{Periwinkle!10}\cmark & \cellcolor{Periwinkle!10}\textbf{41.78} & \cellcolor{Periwinkle!10}\textbf{6.39} & \cellcolor{Periwinkle!10}\textbf{0.56} & \cellcolor{Periwinkle!10}\textbf{0.66} & \cellcolor{Periwinkle!10}\textbf{0.62} & \cellcolor{Periwinkle!10}\textbf{0.85} \\ 
        \bottomrule
    \end{tabular}
    }
    \vspace{0.1cm}
    \caption{Ablation study of the \floss components. (Note: The Precision, Recall, Density and Coverage are provided for unguided settings)}
    \label{tab:focal}
\end{table}

\paragraph{Frequency-term and scheduler ablation} The frequency-to-pixel schedule of Equation~\ref{eq:vfloss} is agnostic to the specific frequency-domain loss used. To test the effectiveness of this schedule, we replace \floss with alternative frequency objectives in the scheduled loss. Table~\ref{tab:schedule_ablation} reports the results using JiT-B at 256² resolution. We compare a discrete Focal Frequency Loss (row 2), the original loss of~\cite{ffl} (row 3), and a Laplacian pyramid loss (row 4). While the schedule proves beneficial across all variants by Epoch 320, the alternative frequency losses perform substantially worse than \floss and \vloss early in training (Epoch 80), yielding FIDs of 21.22 and 23.93 for the loss of~\cite{ffl} and the Laplacian loss, respectively. Only the discrete Focal Frequency Loss is competitive with \floss early on. However, only the continuous formulation \vfloss yields the best FID at every stage of training.

\begin{table}[h]
    \centering
    \begin{tabular}{lccc}
        \toprule
        \textbf{Loss} & \textbf{Epoch 80} & \textbf{Epoch 200} & \textbf{Epoch 320} \\
        \midrule
        \vlosstablecolor & 8.44 & 4.52 & 4.11 \\
        Discrete Focal Frequency Loss $\to$ \vlosstablecolor & 8.92 & 4.39 & 4.13 \\
        Focal Loss from~\cite{ffl} $\to$ \vlosstablecolor & 21.22 & 6.80 & 4.72 \\
        Laplacian loss $\to$ \vlosstablecolor & 23.93 & 5.04 & 4.12 \\
        \rowcolor{Periwinkle!10}
        \vflosstablecolor & \textbf{7.05} & \textbf{4.20} & \textbf{3.95} \\
        \bottomrule
    \end{tabular}
    \vspace{0.1cm}
    \caption{\textbf{Frequency-to-pixel schedule with alternative frequency losses.} Guided FID (B-size model, $256^2$) when the frequency-to-pixel schedule of Equation~\ref{eq:vfloss} is applied to different frequency-domain losses instead of \floss.}
    \label{tab:schedule_ablation}
\end{table}

\subsection{Frequency-Domain Error Analysis}
\label{app:freqband}

Figure~\ref{fig:wallclock_frequency} (right) and Figure~\ref{fig:freqband} (full) measures the average magnitude error between generated and real images in the low- and high-frequency bands of the Fourier spectrum. In the earliest stages of training (50k-75k steps), \floss achieves a lower error than \vloss in both bands, confirming that it corrects the spectral bias faster. As training progresses, \floss plateaus and is eventually overtaken by \vloss. \vfloss combines the fast early spectral alignment of \floss with the late-stage refinement of \vloss, reaching the lowest error in both bands by the end of training.

$f$-loss forces the model to escape this low-frequency bias by actively equalizing the learning signal across all frequency bands. But it is insensitive to phase, which carries the precise spatial location of edges and textures. \vloss directly supervises pixel values and is ultimately better suited for enforcing strict local spatial coherence and phase alignment (e.g., locking sharp edges into exact pixel locations). We hypothesize that this fundamental difference explains why the f-loss eventually plateaus when used alone, making the transition back to the spatial-domain v-loss necessary for final refinement.

\begin{figure}[]
    \centering
    \resizebox{\textwidth}{!}{
        \begin{tikzpicture}
    \begin{groupplot}[
        group style={
            group size=2 by 1,
            horizontal sep=1.8cm,
        },
        width=7cm,
        height=7cm,
        axis lines=left,
        grid=major,
        grid style={dotted, gray!30},
        xlabel={Training step},
        ylabel={Average magnitude error},
        xtick={100,200,300,400},
        xticklabels={100k,200k,300k,400k},
        every axis plot/.append style={
            line width=2.5pt,
            mark=*,
            mark size=3.5pt,
            mark options={fill=white, thick}
        },
    ]

    \nextgroupplot[
        title={Low frequencies},
        ymin=3, ymax=15,
    ]
        \addplot[color=pastelgreen] coordinates {
            (50, 14.00) (75, 8.87) (100, 6.24) (150, 5.74)
            (200, 4.61) (250, 4.67) (300, 4.51) (400, 4.15)
        };

        \addplot[color=pastelblue] coordinates {
            (50, 12.34) (75, 7.96) (100, 7.07) (150, 6.64)
            (200, 6.26) (250, 6.44) (300, 5.94) (400, 6.03)
        };

        \addplot[color=red] coordinates {
            (50, 12.35) (75, 7.72) (100, 7.03) (150, 6.36)
            (200, 5.35) (250, 4.41) (300, 3.85) (400, 3.79)
        };

    \nextgroupplot[
        title={High frequencies},
        ymin=3.2, ymax=4.6,
    ]
        \addplot[color=pastelgreen] coordinates {
            (50, 4.51) (75, 4.18) (100, 3.83) (150, 3.69)
            (200, 3.50) (250, 3.48) (300, 3.48) (400, 3.38)
        };

        \addplot[color=pastelblue] coordinates {
            (50, 4.45) (75, 4.06) (100, 3.88) (150, 3.76)
            (200, 3.65) (250, 3.62) (300, 3.53) (400, 3.49)
        };

        \addplot[color=red] coordinates {
            (50, 4.46) (75, 4.07) (100, 3.90) (150, 3.72)
            (200, 3.55) (250, 3.45) (300, 3.36) (400, 3.34)
        };

    \end{groupplot}

    \coordinate (legend-anchor) at ($(group c1r1.north)!0.5!(group c2r1.north)$);
    \node[
        above=0.8cm of legend-anchor,
        anchor=south,
        inner sep=3pt,
        draw=gray!50,
        rounded corners=2pt,
    ] (legend) {
        \begin{tikzpicture}
            \begin{axis}[
                hide axis,
                xmin=0, xmax=1, ymin=0, ymax=1,
                legend columns=3,
                legend style={
                    draw=none,
                    fill=none,
                    /tikz/every even column/.append style={column sep=1.5cm},
                    font=\small,
                },
                legend entries={$v$-loss, $f$-loss, $fv$-loss},
            ]
                \addlegendimage{color=pastelgreen, mark=*, line width=2.5pt, mark size=3.5pt, mark options={fill=white, thick}}
                \addlegendimage{color=pastelblue, mark=*, line width=2.5pt, mark size=3.5pt, mark options={fill=white, thick}}
                \addlegendimage{color=red, mark=*, line width=2.5pt, mark size=3.5pt, mark options={fill=white, thick}}
            \end{axis}
        \end{tikzpicture}
    };

\end{tikzpicture}
    }
    \caption{\textbf{Spectral error during training.} Average magnitude error in the low- (left) and high-frequency (right) bands. \floss corrects the spectral bias fastest early on but plateaus, while \vfloss reaches the lowest error in both bands by the end of training.}
    \label{fig:freqband}
\end{figure}

\subsection{Generalization to Latent-Space Models}
\label{app:latent}

The spectral bias we address in this work is predominantly a pixel-space phenomenon. When operating in a compressed latent space, the data has already passed through a VAE, which by construction discards much of the high-frequency detail present in pixel space~\cite{rombach2022high}, so it is unclear whether -- or how -- the same bias manifests in latent space. To probe this empirically, we train an SiT-B~\cite{ma2024sit} model with $x$-prediction (required to compute the Fourier spectrum) using $v$-loss and \floss. Table~\ref{tab:sit_latent} shows that \floss still yields a lower FID than $v$-loss at epoch 80, though neither reaches the performance of the original SiT model trained with $v$-prediction (FID 33 at epoch 80). This suggests that while the mechanism we identify is chiefly a pixel-space one, some benefit of frequency-domain supervision may transfer to latent-space training, which we leave for future work.

\begin{table}[h]
    \centering
    \begin{tabular}{lcc}
        \toprule
        \textbf{Loss} & \textbf{SiT output} & \textbf{FID @ Epoch 80} \\
        \midrule
        \vlosstablecolor & $x$-pred & 41.35 \\
        \flosstablecolor & $x$-pred & \textbf{36.38} \\
        \bottomrule
    \end{tabular}
    \vspace{0.1cm}
    \caption{\textbf{Generalization to the latent-space SiT~\cite{ma2024sit} backbone.} FID comparison between $v$-loss and \floss, both using $x$-prediction. Neither reaches the FID of the original SiT model trained with $v$-prediction (FID 33 at epoch 80).}
    \label{tab:sit_latent}
\end{table}

\section{Toy Example}

\begin{figure}
    \centering
    \resizebox{\textwidth}{!}{
        \input{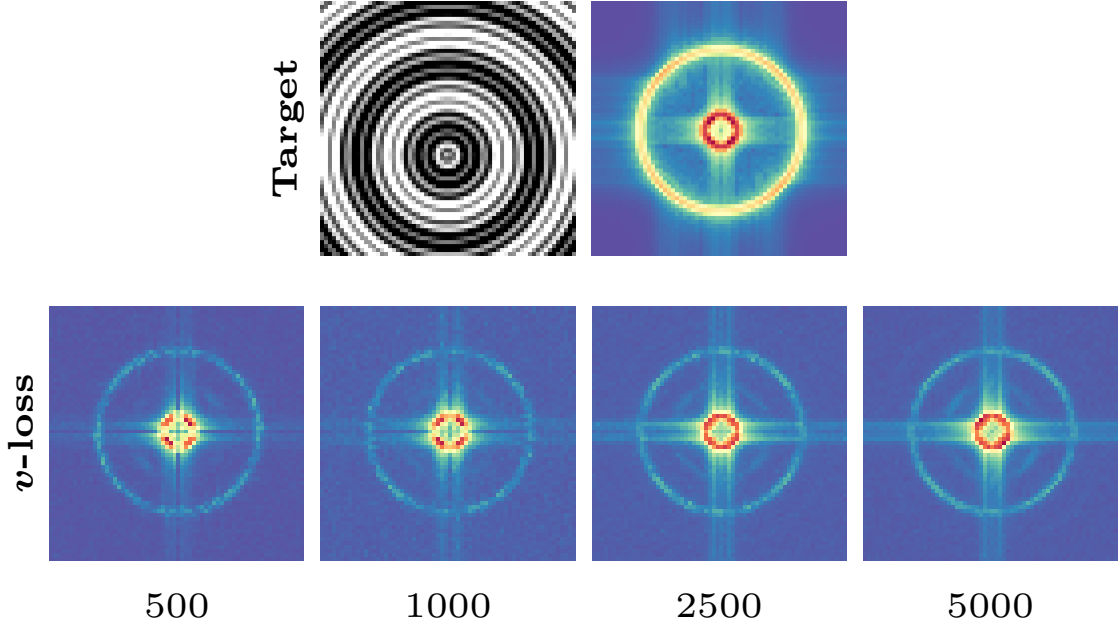}
    }
    \caption{\textbf{Spectrum of the toy experiment of Section~\ref{sec:method}.} (top) Real image and its corresponding spectrum. (bottom) Spectrum of images generated with the MLP at various training steps. The low frequencies (inner ring) are learned much faster than the high frequencies (outer ring).}
    \label{fig:toy_example_spectrum}
\end{figure}

In section~\ref{sec:spectral_diagnosis}, we isolate the temporal dynamics of the spectral bias. We train a small MLP to generate synthetic images composed of exactly two frequency components: one low and one high. In addition to the evolution of the magnitude of the Fourier spectrum in Figure~\ref{fig:spectral_biais}, we provide the raw spectrum in Figure~\ref{fig:toy_example_spectrum}.

\section{Focal Frequency Loss as a Continuous Laplacian Pyramid Loss}
\label{app:laplacian}

\begin{table}
    \centering
    \resizebox{\textwidth}{!}{
    \begin{tabular}{l@{\;\;\;}!{\color{gray!60}\vrule}@{\;\;\;}c@{\;\;\;}!{\color{gray!60}\vrule}@{\;\;}c@{\;\;}!{\color{gray!60}\vrule}@{\;\;}c@{\;\;}!{\color{gray!60}\vrule}@{\;\;}c@{\;\;}!{\color{gray!60}\vrule}@{\;\;}c@{\;\;}!{\color{gray!60}\vrule}@{\;\;}c@{\;\;}!{\color{gray!60}\vrule}@{\;\;}c@{\;\;}}
        \toprule
        \textbf{Loss} & \textbf{\#Bands} & \multicolumn{2}{c@{\;\;}!{\color{gray!60}\vrule}@{\;\;}}{\textbf{FID} {\color{gray!80}$\downarrow$}} & \textbf{Prec.} {\color{gray!80}$\uparrow$} & \textbf{Rec.} {\color{gray!80}$\uparrow$} & \textbf{Den.} {\color{gray!80}$\uparrow$}& \textbf{Cov.} {\color{gray!80}$\uparrow$} \\
        &  (Freq x Orient)  & \footnotesize{cfg=1.0} & \footnotesize{cfg=3.5} \\
        \midrule
        \texttt{\vlosstablecolorttt} & \textcolor{Gray!70}{--} & \textcolor{Gray!70}{49.63} & \textcolor{Gray!70}{7.70} & \textcolor{Gray!70}{0.52} & \textcolor{Gray!70}{0.64} & \textcolor{Gray!70}{0.53} & \textcolor{Gray!70}{0.79} \\
        \greyrule
        \multirow{5}{*}{\flosstablecolorttt} &  $ 2\times 1$ & 45.93 & 8.78 & 0.53 & 0.64 & 0.55 & 0.81 \\ 
        & $2\times 4$ & 47.48 & 7.87 & 0.52& 0.64 & 0.55 & 0.82 \\
        & $3\times 4$ & 44.81 & 6.89 & 0.54 & 0.65 & 0.57 & 0.82 \\
        & $4\times 4$ & 43.79 & 6.63 & 0.55 & 0.65 & 0.59 & 0.83 \\
        \greycmidrule{2-8}
        & \cellcolor{Periwinkle!10}$\infty$  & \cellcolor{Periwinkle!10}\textbf{41.78} & \cellcolor{Periwinkle!10}\textbf{6.39} & \cellcolor{Periwinkle!10}\textbf{0.56} & \cellcolor{Periwinkle!10}\textbf{0.66} & \cellcolor{Periwinkle!10}\textbf{0.62} & \cellcolor{Periwinkle!10}\textbf{0.85} \\ 
        \bottomrule
    \end{tabular}
    }
    \vspace{0.1cm}
    \caption{Comparison between discrete \& continuous Focal Frequency loss. (Note: The Precision, Recall, Density and Coverage are provided for unguided settings) 
    }
    \label{tab:discrete_focal}
    
\end{table}

    

\paragraph{The Laplacian Pyramid}~\cite{burt1983laplacian} decomposes an image into 
bandpass residuals $\{b_0,...,b_{K-1}\}$ and a low-pass base $l_K$. Each level 
captures spatial frequencies within a specific octave, treating scale-specific 
features as independent signals~\cite{heeger1995pyramid}. By performing recursive 
Gaussian blurring and subtraction, it isolates structural residuals at specific 
resolution scales, effectively partitioning the frequency spectrum into octave bands. 
By treating each octave ($2^n$ to $2^{n+1}$) as an independent signal, the pyramid 
elevates sparse, high-frequency details to the same structural importance as the 
low-frequency base — preventing high-energy components from numerically overwhelming 
fine textural details.

\paragraph{Generalizing Laplacian Pyramids} We propose \floss as a continuous, 
infinite-band generalization of octave-based decomposition. By mapping images into 
the frequency domain via the DFT, we gain access to the full spectral hierarchy in 
a single pass. For a frequency at coordinate $(u, v)$, we define:
\begin{equation} 
\mathcal{L}_{u, v} = \log\left(1 + \lVert F_{\theta}(u, v) - F(u, v)\rVert_2\right)
\end{equation}
The spectral difference $e_{u, v} = \lVert F_{\theta}(u, v) - F(u, v)\rVert_2$ 
serves as the functional equivalent of a Laplacian residual. The logarithmic mapping 
linearizes the hierarchical scale of the frequency spectrum: every doubling of 
frequency is granted equal structural weight, preventing the loss from being dominated 
by high-energy, low-frequency components. Our approach thus inherits the perceptual 
benefits of the Laplacian pyramid without the burden of explicit hierarchical 
branching.

\paragraph{Empirical study} Table~\ref{tab:discrete_focal} reports results when 
transitioning from a discrete spectral decomposition to our continuous \floss. We 
partition the Fourier domain into a fixed number of frequency bands ($m=2, 3, 4$) 
and orientations ($k=1, 4$), computing the focal loss independently within each 
cell. Even a coarse discretization already improves over \vlosstablecolorttt, and 
performance consistently improves as the number of spectral cells increases — from 
FID $47.48$ at $2\times4$ bands to $43.79$ at $4\times4$ bands. Our continuous 
formulation achieves the best performance with FID $41.78$, confirming that 
fine-grained spectral weighting provides the most effective way to balance the 
learning signal across the image spectrum.

\section{Extended comparison in early stages of training}

\begin{figure}
    \centering
    \resizebox{\textwidth}{!}{
    \input{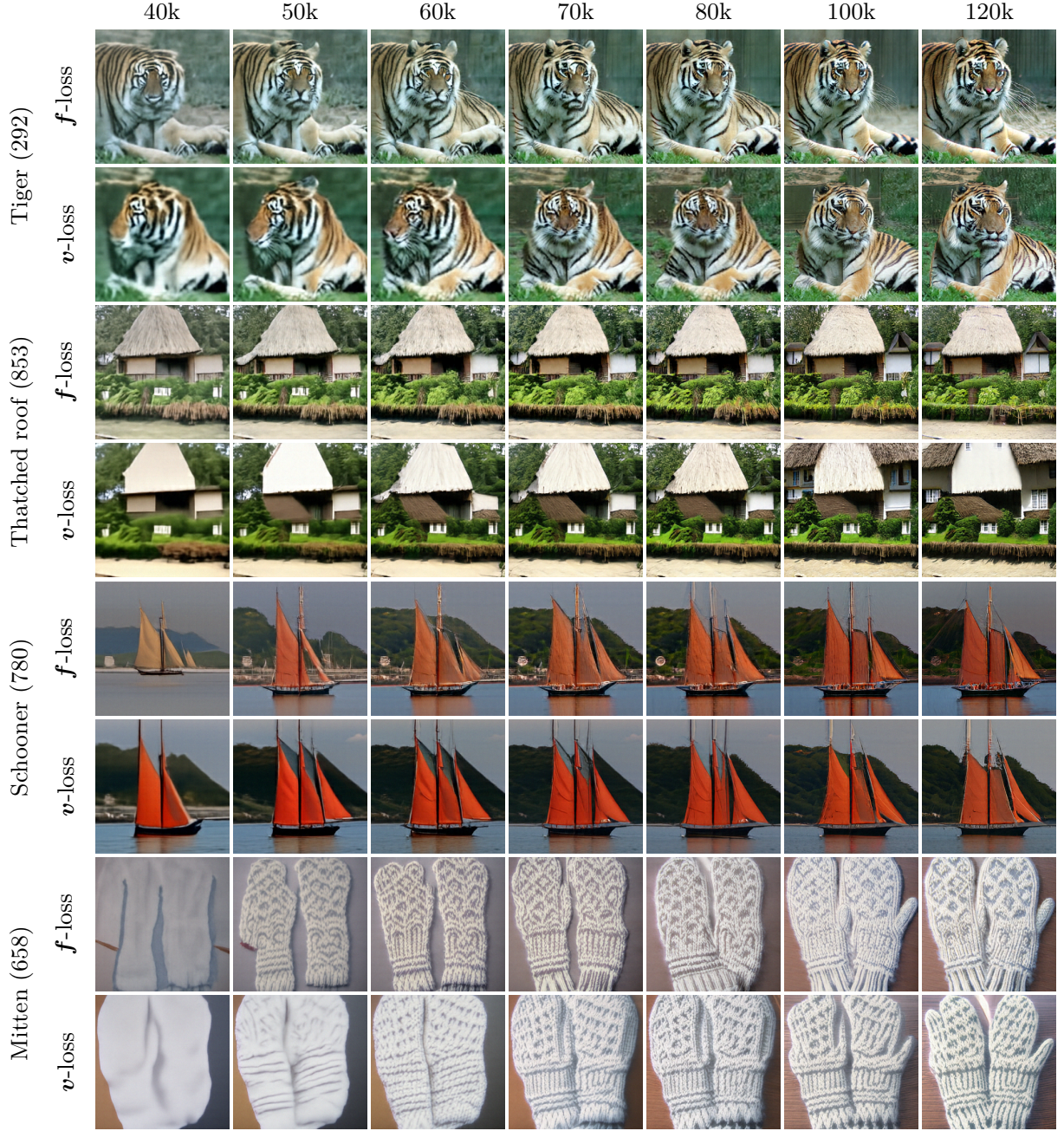}
    }
    \caption{Extended comparison between \floss and \vloss in the early stages of training.}
    \label{fig:extended_evolution}
\end{figure}

Figure~\ref{fig:extended_evolution} shows an extended quantitative comparison between \floss and \vloss in the early stages of training, extanding Figure~\ref{fig:evolution}

\section{Additional qualitative results at $512\times 512$ resolution}

\begin{figure*}[p]
    \centering
    \setlength{\tabcolsep}{2pt} 
    \renewcommand{\arraystretch}{0} 
    \begin{tabular}{cccc}
        \includegraphics[width=0.24\textwidth]{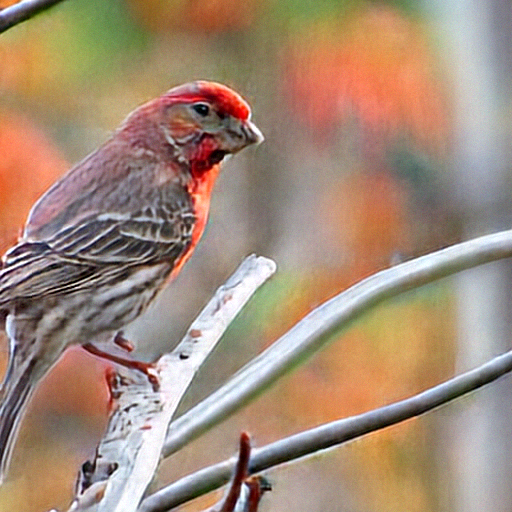} &
        \includegraphics[width=0.24\textwidth]{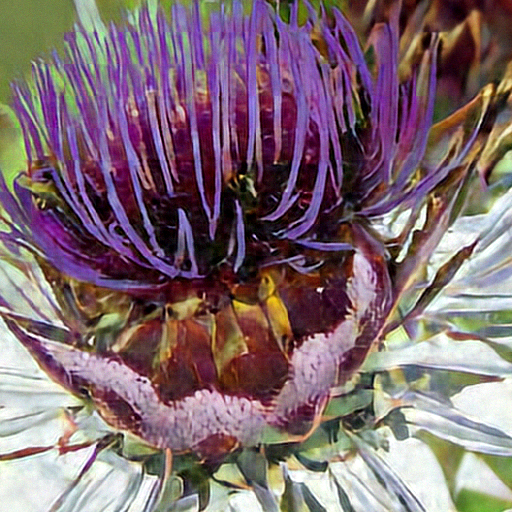} &
        \includegraphics[width=0.24\textwidth]{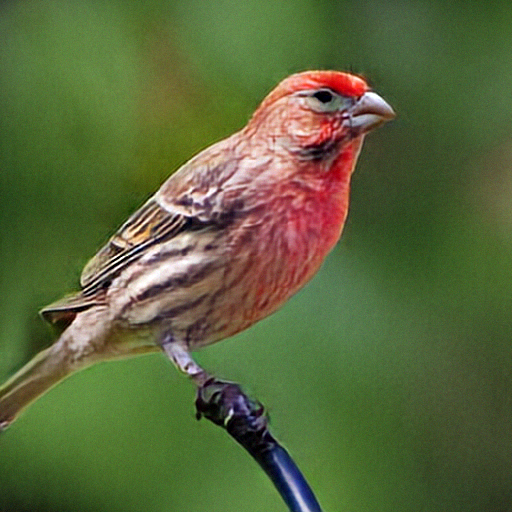} &
        \includegraphics[width=0.24\textwidth]{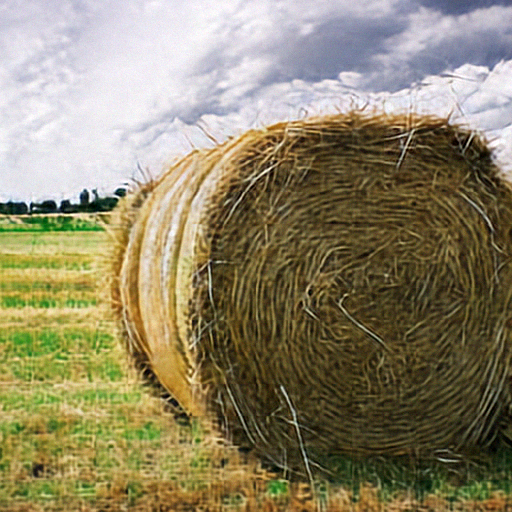} \\ \noalign{\vspace{5pt}}
        \includegraphics[width=0.24\textwidth]{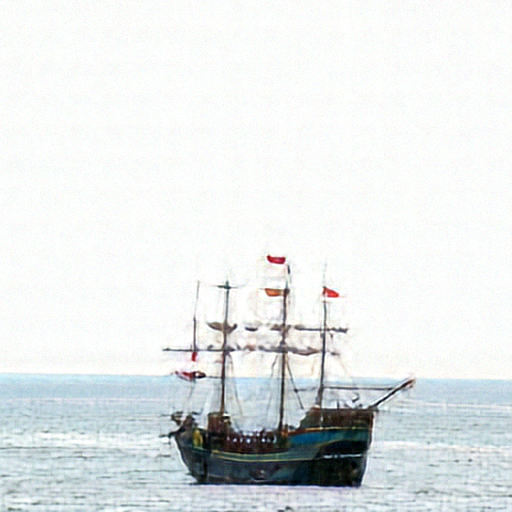} &
        \includegraphics[width=0.24\textwidth]{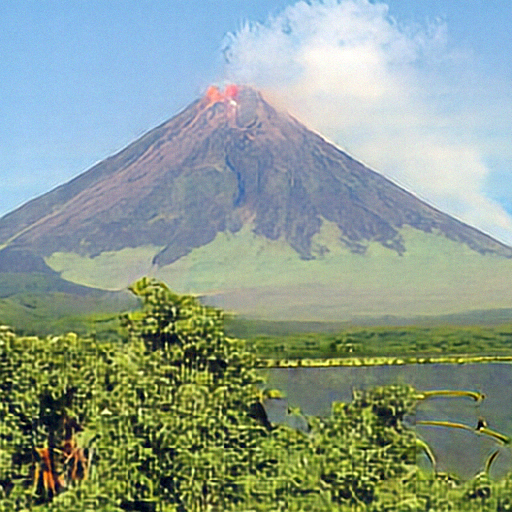} &
        \includegraphics[width=0.24\textwidth]{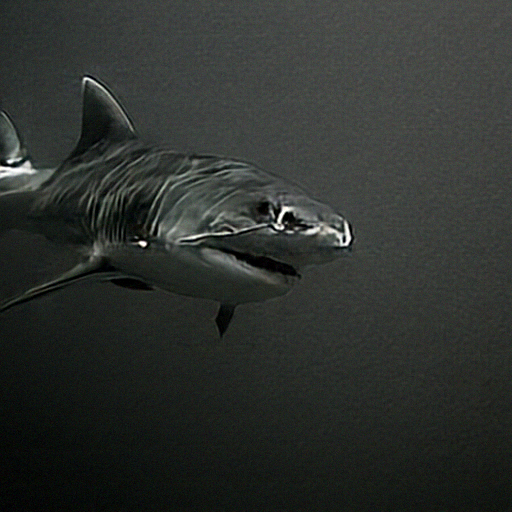} &
        \includegraphics[width=0.24\textwidth]{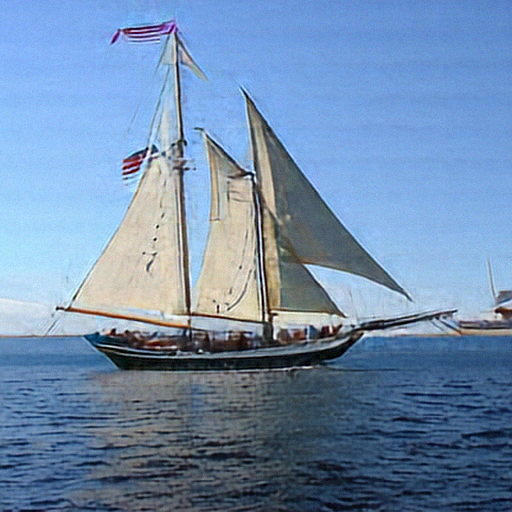} \\ \noalign{\vspace{5pt}}
        \includegraphics[width=0.24\textwidth]{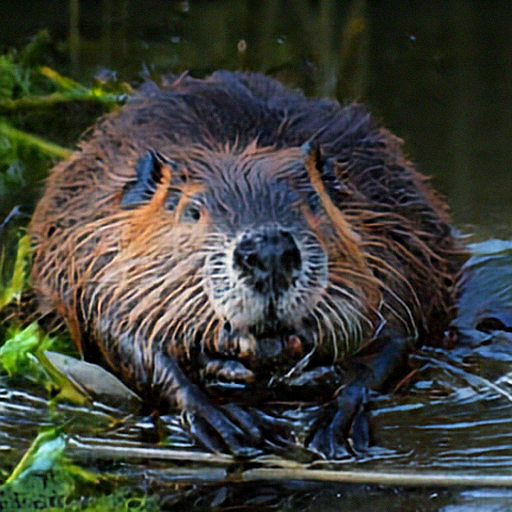} &
        \includegraphics[width=0.24\textwidth]{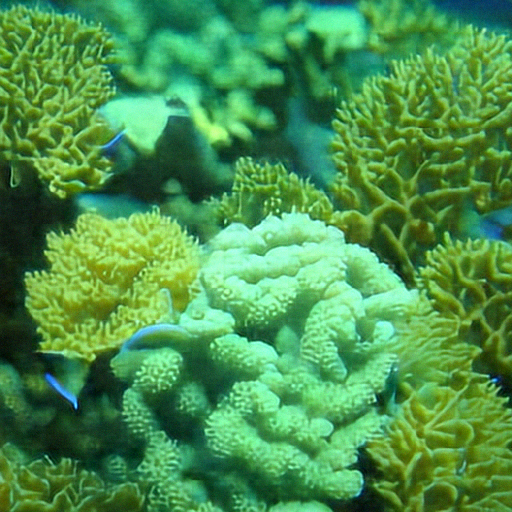} &
        \includegraphics[width=0.24\textwidth]{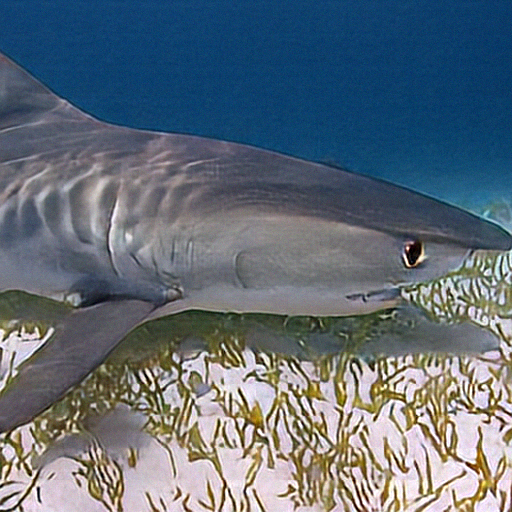} &
        \includegraphics[width=0.24\textwidth]{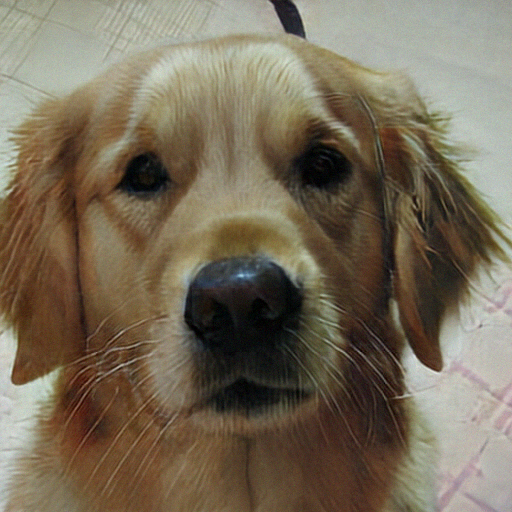} \\ \noalign{\vspace{5pt}}  
    \end{tabular}
    \caption{Additional Qualitative Results at $512^2$ resolution. Model trained with \vfloss for 400k steps. }
    \label{fig:image_grid_512}
\end{figure*}

Figure~\ref{fig:image_grid_512} shows $512^2$ images generated by our model trained with \vfloss.

\end{document}